\documentclass{article}

\PassOptionsToPackage{numbers, compress}{natbib}
\usepackage[preprint]{neurips_2024}

\usepackage[utf8]{inputenc} 
\usepackage[T1]{fontenc}    
\usepackage{hyperref}       
\usepackage{url}            
\usepackage{booktabs}       
\usepackage{amsfonts}       
\usepackage{nicefrac}       
\usepackage{microtype}      
\usepackage{xcolor}         
\usepackage{booktabs}

\usepackage{enumitem}
\usepackage{amsmath}
\usepackage{hyperref} 
\usepackage[capitalise]{cleveref} 
\usepackage{tabularx}
\crefname{section}{\S}{\S\S}

\usepackage{diagbox}
\usepackage{wrapfig}
\usepackage{amssymb}

\usepackage{array, makecell}
\usepackage{amsmath}
\usepackage{multirow}
\usepackage{float}
\usepackage{algorithm}
\usepackage{tikz}
\usepackage{pgfmath}
\usepackage{tabularx}
\usepackage{algorithmic}
\usepackage{graphicx}
\usepackage{subcaption}
\usepackage{multicol}

\usepackage{amssymb}
\usepackage{mathtools}
\usepackage{amsthm}
\usepackage{bbm}
\usepackage{ifthen}

\newtheorem{theorem}{Theorem}
\newtheorem{remark}{Remark}
\newtheorem{proposition}{Proposition}

\newcommand{\smartparagraph}[1]{\noindent{\bf #1}\ }

\DeclareMathOperator*{\argmin}{arg\,min}
\DeclareMathOperator{\GGC}{GGC}
\DeclareMathOperator{\BGGC}{BGGC}

\newif\ifrevision
\revisiontrue

\ifrevision

\else

\fi

\usepackage{xcolor}

\title{Decentralized Personalized Federated Learning}

\author{%
  Salma Kharrat\\ 
  Computer, Electrical, and Mathematical Sciences and Engineering Division \\
  King Abdullah University of Science and Technology (KAUST)\\
  \texttt{salma.kharrat@kaust.edu.sa} \\
  \AND
  Marco Canini\\ 
  Computer, Electrical, and Mathematical Sciences and Engineering Division\\
  King Abdullah University of Science and Technology (KAUST)\\
  \AND
  Samuel Horvath\\ 
   Machine Learning Department\\
  Mohamed bin Zayed University of Artificial Intelligence (MBZUAI)
\\
}

\begin{document}
\maketitle
\begin{abstract}
This work tackles the challenges of data heterogeneity and communication limitations in decentralized federated learning. We focus on creating a collaboration graph that guides each client in selecting suitable collaborators for training personalized models that leverage their local data effectively. Our approach addresses these issues through a novel, communication-efficient strategy that enhances resource efficiency. Unlike traditional methods, our formulation identifies collaborators at a granular level by considering combinatorial relations of clients, enhancing personalization while minimizing communication overhead. We achieve this through a bi-level optimization framework that employs a constrained greedy algorithm, resulting in a resource-efficient collaboration graph for personalized learning. Extensive evaluation against various baselines across diverse datasets demonstrates the superiority of our method, named DPFL. DPFL consistently outperforms other approaches, showcasing its effectiveness in handling real-world data heterogeneity, minimizing communication overhead, enhancing resource efficiency, and building personalized models in decentralized federated learning scenarios.
\end{abstract}

\vspace{-0.2cm}
\section{Introduction}
\vspace{-0.1cm}
The ongoing unprecedented growth in data captured and stored on edge devices has led to a flurry of research proposing collaborative learning paradigms that do not necessitate transferring data to a centralized location, thereby preserving the privacy of the data. A popular approach is federated learning (FL)~\cite{mcmahan2017communication}, where clients share model parameters instead of data samples. In this approach, training occurs over a series of rounds, alternating between a phase of training on client devices using their locally-held data followed by an aggregation phase of the clients' model updates.

A key challenge in FL is the presence of statistical heterogeneity, whereby models are trained with clients' data that are not independent and identically distributed (non-IID)~\cite{kairouz2021advances}.
This heterogeneity may lead to local models drifting apart, making it challenging to derive a globally satisfactory model~\cite{subopt, fourati2023filfl}. Moreover, real-world scenarios frequently lack prior knowledge of data distributions, making it challenging to quantify data heterogeneity beforehand.

Catering to the inherent heterogeneity, personalized FL approaches (e.g., \cite{apfl, ditto, pfedme, hanzely2020federated, fedrep})  aim to train for each client, a model tailored to its specific data patterns.
In this work, we advocate for targeted collaboration, identifying \emph{beneficial collaborators for each client}. In this scenario, decentralized learning becomes particularly relevant since communication can be restricted to collaborating clients, thereby reducing communication overhead.

Indeed, decentralized learning -- where clients directly collaborate within a peer-to-peer network -- has recently witnessed a surge in interest by researchers~\cite{duchi2011dual, wei2012distributed, colin2016gossip, lian2017can, lian2018asynchronous,  jiang2017collaborative, tang2018d, vanhaesebrouck2017decentralized, bellet2018personalized}.
In this paradigm, devices form a sparse collaboration graph over which model updates are shared, and there isn't a central server to orchestrate the process, enhancing resilience against single points of failure while also offering enhanced privacy protection~\cite{cyffers2022privacy}, allowing for faster model training~\cite{lian2017can}, and providing robustness towards slow client devices~\cite{neglia2019role}.

However, a crucial step in decentralized learning is the need to \emph{identify the collaboration graph}, i.e., identify for each client its suitable collaborators. The quality of the collaboration graph significantly impacts the performance of the resulting personalized models, which determines the effectiveness of collaboration among clients. We note that there is a strong interdependence between data heterogeneity and collaboration. For instance, if data distributions among clients are relatively homogeneous, collaborative learning will likely benefit every client involved. However, as data heterogeneity increases, ``blind'' collaboration may have negative consequences for the performance of some or all clients.
It is challenging to construct an optimal collaboration graph (e.g., one that maximizes clients' model performance) because it must account for two key factors: data heterogeneity and resource constraints (network bandwidth, memory, and compute resources).
This paper seeks to devise a solution to the above challenges.

Several methods have been proposed to personalize models for better adaptation to client data, utilizing multi-task learning~\cite{smith2017federated}, meta-learning~\cite{jiang2019improving, fallah2020personalized, al2021data}, transfer learning~\cite{wang2019federated, mansour2020three}, and knowledge distillation~\cite{li2019fedmd, chen2023spectral}. However, these methods rely on a global model. They handle personalization by either including a regularization term to encourage the local model to stay close to the global model, considering an interpolation between local and global models, or fine-tuning global models to achieve personalized solutions. These approaches restrict the scalability of these methods in decentralized settings. Moreover, they disregard the importance of identifying, in a fine-grained manner and from each client's perspective, the set of collaborating devices that can best enhance its model.

In this work, we propose a novel bi-level optimization problem aiming at jointly optimizing the clients' models and the collaboration graph while ensuring communication and resource efficiency. The driving principle is that each client will collaborate only with clients that yield positive returns for their collaboration. We define this positive return through a reward function, assessing the cooperation of a set of clients. The algorithm we propose for graph construction leverages the combinatorial effect of clients while also restricting the number of nodes a client can communicate with and the number of models it can store due to memory constraints. This further optimizes communication and resource efficiency. According to our formulation, the aggregated model for each client is directly determined by the inward edges connecting that client to others. Our method can infer beneficial collaborators for each client, effectively addressing data heterogeneity without requiring any prior knowledge about the data distribution.

It is crucial to note that unlike many prior state-of-the-art solutions (e.g., \cite{ye2023personalized, li2022learning, li2022towards}), our collaboration graph doesn't necessarily have to be symmetric; that is, client $\textit{A}$ can benefit from collaborating with client $\textit{B}$, while client $\textit{B}$ may receive a negative reward if it collaborates with client $\textit{A}$. This situation could arise, for example, in a scenario where client $\textit{B}$ has a large number of data samples, and the optimal strategy for it might be to collaborate with no one. Conversely, other clients with similar data distributions but smaller datasets might find collaboration with client $\textit{B}$ highly valuable. 

Beyond our main objective of finding more fine-grained collaborators, which can be achieved through potentially directed edges, allowing for directed graphs offers several advantages over undirected graphs. First, a directed graph is a more general and relaxed assumption than an undirected graph, as the latter can be viewed as a special case. Moreover, it has been demonstrated that undirected graphs are more susceptible to deadlock in practical scenarios~\cite{tsianos2012consensus, lian2018asynchronous}. Finally, many applications necessitate directed communication networks~\cite{zhang2019fully, nedic2018network}.
\newpage
\smartparagraph{Contributions.} 
\begin{itemize}[leftmargin=1.0cm]
\item We formulate a novel decentralized personalized FL problem, which introduces a constrained discrete combinatorial objective within a bi-level optimization framework. This framework jointly accounts for graph generation, model personalization, and resource constraints. 
    \item We propose a DPFL algorithm, which alternates between optimizing personalized models and inferring sets of collaborators through an efficient approximation algorithm. Graph optimization relies on the marginal gains of adding or removing collaborators, thereby considering the combinatorial effects of clients instead of relying on pairwise collaborator selection.
    \item We devise a solution for asymmetric graph construction, offering enhanced granularity for collaborators' selection and improved communication and resource efficiency.
    \item We conduct extensive experiments on various datasets under diverse, realistic scenarios and evaluate our approach against eleven methods between personalized FL methods and other baselines; our empirical results demonstrate that DPFL offers superior performance.
\end{itemize}

\section{Methodology}
We introduce a novel formulation for the optimization problem in personalized FL. First, we provide the underlying intuition behind the problem and present the comprehensive problem formulation in \cref{sec:full_problem_formulation}. Next, we propose decomposing the problem and employing an alternating minimization \cref{sec:Problem_decomposition}.

\subsection{Optimization problem}
\label{sec:full_problem_formulation}
Generalization in FL aims to fit a single global model with parameters $\mathbf{w}$ to the non-IID data held by individual clients \cite{mcmahan2017communication}. This distributed optimization problem can be represented as follows:
\begin{equation}
\label{eq:FL}
  \min _{\mathbf{w} \in \mathbb{R}^d}\left\{F^{\mathcal{D}}(\mathbf{w}) \triangleq \sum_{k=1}^N p_k F_k(\mathbf{w})\right\},
\end{equation}
where $N$ is the number of devices, $\mathcal{D}$ is a global distribution, $p_k \geq 0$ is the weight of the $k$-th device s.t. $\sum_{k=1}^N p_k=1$. It is common to select $p_k$ proportional to device $k$'s dataset size. Suppose the $k$-th device holds data drawn from the distribution $\mathcal{D}_k$. The local objective $F_k(\cdot)$ is defined as:
$
F_k(\mathbf{w}) \triangleq \mathbb{E}_{x_{k} \sim \mathcal{D}_k} \ell\left(\mathbf{w} ; x_{k}\right)
$
where $\ell(\cdot ; \cdot)$ is some loss function.

In personalized FL, the primary objective shifts from seeking a single global model that can generalize across all clients to the pursuit of local models tailored to each client. These local models are designed to perform well on unseen data samples drawn from the same distribution as the client's local training data. Consequently, the original objective function in \cref{eq:FL} becomes the following:
\begin{equation}
\label{eq:perso}
\min _{\mathbf{\mathbf{w}} = \{\mathbf{w}_i  \in R^{d}\}_{i=1}^N}\left\{F(\mathbf{\mathbf{w}}) \triangleq  \sum_{k=1}^N p_k F_k\left(\mathbf{w}_k\right)\right\}. 
\end{equation}

Considering significant heterogeneity in the data distributions among the clients, an intuitive solution to the problem outlined in  \cref{eq:perso} is that each client independently trains its local model using its local data, obviating the necessity for collaboration with other clients. However, this holds only if each client has access to an infinite number of IID samples from its local distribution $\mathcal{D}_k$ or the local distributions are significantly different.

We focus on collaborative learning in real-world settings, where limited data access and heterogeneity are critical challenges. Clients in such scenarios operate with a finite number of local samples.
While we advocate for inter-client collaboration, this collaboration must align with, not compromise, our core objective of personalized model effectiveness. Collaboration needs to be designed to enhance personalized model performance within this complex data landscape while respecting communication and resource constraints, particularly in decentralized settings where coupled all-to-all communication is prohibitive. Therefore, accounting for the application-imposed resource budget becomes crucial. 

We reformulate our objective function to address these challenges and optimize model performance within this dynamic environment. Thus, we consider this problem formulation:
\begin{equation}
\label{eq:perso_collaborators_initial}
\min_{\substack{\mathbf{\mathbf{w}} = \{\mathbf{w}_i \in \mathbb{R}^{d} \}_{i=1}^N\\
\mathbf{\mathcal{C}} = \{\mathcal{C}_i \in 2^{[N] \setminus \{i\}};\textbf{  } |\mathcal{C}_i| \leq B_c\}_{i=1}^N,\\ 
 }} 
\left\{F(\mathbf{\mathbf{w}},\mathbf{\mathcal{C}}) \triangleq \sum_{k=1}^N p_k F_k\left(\mathbf{w}_k, \mathcal{C}_k\right)\right\}. \\
\end{equation}

In this optimization problem (\cref{eq:perso_collaborators_initial}), our objective is to minimize the function $F(\mathbf{\mathbf{w}}, \mathbf{\mathcal{C}})$, where $\mathbf{\mathbf{w}}=\left\{\mathbf{w}_1, \ldots, \mathbf{w}_N\right\}$ represents a set of parameters and $\mathbf{\mathcal{C}}=\left\{\mathcal{C}_1, \ldots, \mathcal{C}_N\right\}$ denotes collaboration assignments. $\mathcal{C}_i \cup \{i\}$ contains $j$, if client $i$ receives updates from client $j$. Moreover, in our algorithm design, we enforce $ \forall k \in [N]: |\mathcal{C}_i| \leq B_c$, where $B_c$ represents the resource budget to improve communication and resource efficiency. 
The function $F_k\left(\mathbf{w}_k, \mathcal{C}_k\right)$ is defined as the average loss over a dataset $\mathcal{D}_k$ for client $k$, expressed as:
$$
F_k(\mathbf{\mathbf{w}}_k, \mathcal{C}_k) \triangleq \mathbb{E}_{x_{k} \sim \mathcal{D}_k} \ell\left(\mathbf{w}_k, \mathcal{C}_k ; x_{k}\right).
$$
Here, $\ell\left(\mathbf{w}_k, \mathcal{C}_k, x_{k}\right)$ represents the loss function applied to the parameters $\mathbf{w}_k$, collaboration assignments $\mathcal{C}_k$, and data point $x_{k}$. We can rewrite it as follows: $\ell\left(\mathbf{w}_k,\mathcal{C}_k; x_{k}\right) = \ell( \hat{\mathbf{w}}_k; x_{k})$ where the crucial element in this formulation is $\hat{\mathbf{w}}_k$, computed as the average of individual parameters $\mathbf{w}_i$ within the collaboration set $\mathcal{C}_k$ for client $k$:
\begin{equation}
  \label{eq:avg}
\hat{\mathbf{w}}_k \triangleq \frac{1}{\sum_{i \in \Tilde{\mathcal{C}}_k} p_i} \sum_{i \in \Tilde{\mathcal{C}}_k} p_i\mathbf{w}_i; \text{ where } \Tilde{\mathcal{C}}_k \triangleq \mathcal{C}_k \cup \{k\}.
\end{equation}

\vspace{-0.3cm}
\subsection{Alternating minimization}
\label{sec:Problem_decomposition}
Solving \cref{eq:perso_collaborators_initial} involves concurrently optimizing for both $\mathbf{w}$ and $\mathcal{C}$.
To simplify this complex problem, we propose an alternating minimization approach. In the first sub-problem, we use the given collaboration graph $\mathbf{\mathcal{C}^\star}$ to update local parameters $\mathbf{w}_k$, while in the second sub-problem, we focus on identifying the best set of collaborators $\mathcal{C}_k$ given $\mathbf{w}$.
The first sub-problem can be written as:
\begin{equation}
\label{eq:decomp1}
\min _{{\mathbf{\mathbf{w}} = \{\mathbf{w}_i  \in R^{d}\}_{i=1}^N}}\left\{F(\mathbf{\mathbf{w}},\mathbf{\mathcal{C}^\star}) \triangleq \sum_{k=1}^N p_kF_k\left(\mathbf{w}_k, \mathcal{C}^\star_k\right)\right\}. 
\end{equation}
We propose to solve \cref{eq:decomp1} in a decentralized setting, where each client first performs $\tau$ local updates of their local parameters $\mathbf{w}_k$ based on their local data. This is followed by the aggregation step, in which each client updates their local solution $\mathbf{w}_k$ as defined in \cref{eq:avg}.

For finding the collaboration graph, given fixed $\mathbf{w}^\star$, the collaboration assignment $\mathcal{C}_k$ for each client $k \in [N]$ is found as a solution to the following minimization problem:
\begin{equation}
\label{eq:greedy_eq_t}
 \min_{\mathcal{C}_k \in 
 \Omega_k} F_k^{\mathcal{V}}\left(\frac{1}{\sum_{i \in \mathcal{C}_k \cup \{k\}} p_i} \sum_{i \in \mathcal{C}_k \cup \{k\}} p_i \mathbf{w}_i^\star\right),\\
\end{equation}
where $F_k^{\mathcal{V}}(\cdot)$ represents the validation loss of client $k$ and $\Omega_k \subseteq 2^{[N] \setminus \{k\}}$ such that $|\Omega_k| \leq B_c$ is a set of clients that client $k$ can potentially collaborate with other than himself. Further details about the proposed methods are provided in \cref{sec:algo}.

\vspace{-0.1cm}
\section{Proposed Method}
\label{sec:algo}
\vspace{-0.1cm}
We introduce DPFL, which incorporates the identification of beneficial collaborators alongside FL training. To achieve this, we define a combinatorial objective function on the discrete and large space of client combinations in \cref{eq:greedy_eq_t}, and introduce a constrained greedy algorithm denoted as $\GGC$. This algorithm approximates a solution for this objective by optimizing client combinations to enhance decentralized personalized FL under strict communication and resource constraints.
\subsection{Decentralized Personalized FL (DPFL)}
The primary goal for each client is to identify potential collaborators who can assist in achieving better personalization while adhering to a resource budget constraint, denoted as $B_c$. \cref{alg:DPFL}, DPFL, details the pseudocode of our employed approach. As a preprocessing step, we establish the initial collaboration graph, $\Omega_k$, ensuring adherence to the budget constraint.
 This process commences by initializing each local model as $\mathbf{w}$, followed by conducting $\tau_{\text{init}}$ local training epochs. Subsequently, after each client acquires its initial solution $\mathbf{w}_k^{init}$, it proceeds to determine its collaboration set, $\Omega_k$, as described in \cref{sec:greedy_alg} and \cref{alg:BGGC} in \cref{appendix:efficient_calls}. With the initial graph in place, the main training loop begins, spanning $T$ communication rounds.  During each round, clients perform local training for $\tau_{\text{train}}$ epochs and download the local models $\mathbf{w}_i$ from all clients in their respective collaboration sets $\Omega_k$ where $\Omega_k$ contains at most $B_c$ clients. Upon receiving these models, each client determines 
 
\begin{minipage}{0.42\textwidth} 
its best collaborators for that specific round using \cref{alg:GGC}. Specifically, each client $k$ calculates the weighted average of the models received from $\Omega_k$ that offers the best benefits by minimizing its local validation loss. The resulting set from this step for each client $k$ is $\mathcal{C}_k$, where $|\mathcal{C}_k| \leq |\Omega_k|$ and $\mathcal{C}_k \subseteq \Omega_k$.  In essence, the offline-constructed graph during preprocessing using $\BGGC$ remains static, and the training method serves as an averaging mechanism where the optimal set of collaborators is determined using \cref{alg:GGC}. To reduce the complexity of the graph construction step, we avoid updating the graph structure by removing edges 
\end{minipage}
\hfill
\begin{minipage}{0.56\textwidth}
\vspace{-0.4cm}
\begin{algorithm}[H]
\small
\caption{Decentralized Personalized FL (DPFL)}\label{alg:DPFL}
\begin{algorithmic}[1]
\REQUIRE{$T$, $\mathbf{w}$, $N$, $m$, $B_c$, $\tau_{\text{init}}$, $\tau_\text{train}$}, LocalOpt \\
\FOR[\textsf{Preprocess}]{device $k \in [N]$ {\bf in parallel}} 
    \STATE $\mathbf{w}_k^{init} \leftarrow \text{LocalOpt}(\mathbf{w}, \tau_{\text{init}})$ 
    \STATE $\Omega_k \leftarrow \BGGC(k, [N], m, B_c) \quad$ \COMMENT{Explained in \cref{sec:BGGC}} 
    \STATE $\mathbf{w}_k \leftarrow \frac{1}{\sum_{i \in \Omega_k \cup {k}} p_i} \sum_{i \in \Omega_k \cup {k}} p_i \mathbf{w}_i^{init}$
\ENDFOR
\FOR[\textsf{Training Loop}]{$t= 1, \cdots, T-1$}
    \FOR {device $k \in [N]$ {\bf in parallel}}
    \STATE $\mathbf{w}_k \leftarrow \text{LocalOpt}(\mathbf{w}_k, \tau_{\text{train}})$
    \STATE send $\mathbf{w}_k $, $\forall$ $j \text{ s.t. } k \in \Omega_j$ and receive  $\mathbf{w}_j$, $\forall j \in \Omega_k$,\\
    \STATE $\mathcal{C}_{k} \leftarrow \GGC(k, \Omega_k, m, B_c) \quad$ \COMMENT{Explained in \cref{sec:greedy_alg}} 
    \STATE $\mathbf{w}_k \leftarrow \frac{1}{\sum_{i \in \mathcal{C}_k \cup {k}} p_i} \sum_{i \in \mathcal{C}_k \cup {k}} p_i \mathbf{w}_i$
    \ENDFOR
\ENDFOR
\end{algorithmic}
\end{algorithm}
\end{minipage}

 when GGC does not select a client for aggregation in line 10. In fact, a client might be skipped in the current round due to factors like noise but still may offer benefits in future rounds. In our experiments, local SGD is used as the local optimizer, though DPFL is adaptable to any local optimizer.

\subsection{Greedy Graph Construction (GGC)}
\label{sec:greedy_alg}
Our primary goal is to solve \cref{eq:greedy_eq_t}. In this combinatorial objective, 
client $k$ seeks a set $\mathcal{C}_k$ wherein collaboration with its elements results in a positive return. In our context, this positive return is quantified by a reward function representing the validation loss of each client. This conceptualization is pivotal as our approach departs from previous works, which typically focused on pairs of clients, e.g., \cite{ye2023personalized}, overlooking the significance and synergy of the entire group. For instance, clients $A$ and $B$ collaborating alone might not be ideal, but adding client $C$ to the collaboration set could significantly alter the outcome. In such a scenario, the collaborative efforts of clients $A$, $B$, and $C$ contribute to a decrease in the overall loss experienced by client $A$. We refer to the motivational example in \cref{appendix:motivation} explaining the importance of the combinatorial effect in boosting client performance, in which we demonstrate how relying on pairwise comparisons may fail compared to group synergy.

Finding the optimal set $\mathcal{C}_k$ for each client $k$ requires extensive computational exploration. Furthermore, notice that our objective function in \cref{eq:greedy_eq_t} is not monotonic,\footnote{A set function $f: 2^{\Omega} \rightarrow \mathbb{R}$ is monotone if for any $A \subseteq B \subseteq \Omega$ we have $f(A) \leq f(B)$.} i.e., adding clients to the set $\mathcal{C}_k$ does not always result in a better reward.  To address this, we adapt the non-monotone combinatorial bandit algorithm of \cite{fourati2023randomized}, proposing a greedy graph construction (GGC), detailed in \cref{alg:GGC} in \cref{app:GGC}, which efficiently selects the set $\mathcal{C}_k$. 
The complexity of the employed algorithm is $\mathcal{O}(|\Omega_k|)$, which is at most $\mathcal{O}(B_c)$.

While our objective is to minimize the local validation loss of clients as defined in \cref{eq:greedy_eq_t}, following the literature on combinatorial maximization, it is more common to maximize a reward function. Thus, we define our reward function as follows:
 \vspace{-0.1cm}
 \begin{align}
 \label{eq:reward}
    \mathcal{R}(\mathcal{S})  \triangleq - F_k^{\mathcal{V}}\left(\frac{1}{\sum_{i \in \mathcal{S} \cup \{k\}} p_i} \sum_{i \in \mathcal{S} \cup \{k\}} p_i \mathbf{w}_i\right),
 \end{align}
 Given the defined reward, GGC operates through a series of iterations, each involving a decision-making process for adding or removing clients from two sets, $\mathcal{X}$ and $\mathcal{Y}$. The set $\mathcal{X}$ represents the set of collaborators and initially contains client $k$ which is the client running GGC, and $\mathcal{Y}$ contains clients in $\Omega_k \cup \{$k$\}$. In each step, the algorithm computes two variables, $a$ and $b$, representing the expected marginal gains from adding and removing a specific client $j \in \mathcal{S}$, respectively, and defined as follows: 
$
\begin{aligned}
    &a = \max(\mathcal{R}(\mathcal{X} \cup \left\{j\right\}) - \mathcal{R}(\mathcal{X}), 0) \text{ and } b =  \max(\mathcal{R}(\mathcal{Y} \setminus \left\{j\right\}) - \mathcal{R}(\mathcal{Y}), 0).
\end{aligned}\label{eq:abdefn}
$
Client $k$, running GGC, adds a client $j$ to the set of collaborators, $\mathcal{X}$, greedily with a probability $p = \frac{a}{a + b}$, i.e, a client is likely to be added if the marginal gain of adding it is higher than the marginal gain of removing it ($a \geq b$). If the marginal gain of adding is positive and the marginal gain of removing is negative, then a client is added with $p=1$. Conversely, if the marginal gain of removing is positive and the marginal gain of adding is negative, the client is removed with $p=1$. In cases where both $a$ and $b$ are zero, the algorithm defaults to setting $p = 1$. The process iterates until decisions are made for all individual clients or the cardinality of the set of collaborators $\mathcal{X}$ reaches the budget constraint $B_c$, ultimately identifying the optimized set of clients satisfying the imposed budget constraint.

\smartparagraph{GGC algorithm complexity:}  The overall complexity of $\GGC$ (\cref{alg:GGC}) is $\mathcal{O}(B_c)$, which reduces to constant complexity $\mathcal{O}(1)$ since during training $B_c$ (budget constraint) is constant. $\GGC$ operates by iterating over the received clients as input. It computes two variables $a$ and $b$, requiring constant number forward passes over the network, which translates to $\mathcal{O}(1)$  complexity. Subsequently, it decides whether to add or remove a client, an operation also achieved in $\mathcal{O}(1)$. Therefore, the algorithm's complexity essentially reduces to looping over the clients, resulting in $\mathcal{O}(B_c)$ during the training which does not increase with a growing number of clients as $B_c$ is a predefined constant. 
\vspace{-0.2cm}
\subsection{Batched Greedy Graph Construction (BGGC)}
\vspace{-0.2cm}
\label{sec:BGGC}
As outlined in \cref{sec:Problem_decomposition}, each client $k$ can only collaborate with a set of clients $\Omega_k$, which is restricted in size (i.e., $|\Omega_k| \leq B_c$) to maintain communication and resource efficiency. Moreover, as depicted in \cref{alg:DPFL}, the initial graph construction is conducted as a preprocessing step where each client trains a model locally for a sufficient number of epochs using the same model initialization parameters, resulting in the client's local solution denoted as $\mathbf{w}_k^{init}$ (lines 1-2). Then in line 3, each client $k$ determine its set of beneficial collaborators $\Omega_k$ by minimizing the following optimization problem: 
\begin{equation}
\label{eq:preprocessing}
\min_{\mathcal{S} \subseteq [N]\setminus {k}} F_k^{\mathcal{V}}\left(\frac{1}{\sum_{i \in \mathcal{S} \cup {k}} p_i} \sum_{i \in \mathcal{S} \cup {k}} p_i \mathbf{w}_i^{init}\right) \; \text{s.t.} \; |\mathcal{S}| \leq B_c.
\end{equation}

Although our proposed \cref{alg:GGC} in \cref{sec:greedy_alg} efficiently approximates the solution to our objective \cref{eq:preprocessing} while demonstrating resilience to noisy rewards (c.f. Corollary 2 in \cite{fourati2023randomized}), this method relies on reward computations to the objective in \cref{eq:preprocessing}, necessitating model updates from potential collaborators. This poses a significant challenge when applied to the initial graph generation of $\Omega_k$, where the number of required models may surpass our budget constraint $B_c$, dictated by practical limitations such as network bandwidth, compute resources, and storage. To address this challenge, we propose an enhancement to the algorithm $\GGC$ called $\BGGC$ (Batched Greedy Graph Construction) by streamlining reward computations into efficient batches of communication and computation, each requiring only $\mathcal{O}(B_c)$ resources. Details of this approach are outlined in \cref{alg:BGGC}; see \cref{appendix:efficient_calls}. Note that $\BGGC$ is necessary only during the preprocessing step.  During the training loop (lines 6-12 in \cref{alg:DPFL}), the reward computations of $\GGC$ will require, at most, $B_c$ models, which falls within the imposed budget constraint.

\smartparagraph{BGGC algorithm complexity.} The overall computation complexity of $\BGGC$ is $\mathcal{O}(N)$. However, the communication and resource complexity is only $\mathcal{O}(B_c)$ for each of the $\lceil N/B_c \rceil$ steps. 

\begin{theorem}
\label{theorem1}
Assuming seeded randomness, executing algorithms $\GGC$ and $\BGGC$ with the same seed produces identical results for a given client $k$, clients set $\mathcal{S}$, and budget $B_c$.
\end{theorem}
\vspace{-0.1cm}
The proof of \cref{theorem1} is reported in \cref{app:proof}.
\begin{remark}
\textbf{Differences between GGC and BGGC.} 
 Executing GGC during preprocessing may violate the budget constraint, as the reward computation requires access to weights from all clients. BGGC addresses this issue by adhering to communication and resource constraints, though it requires an additional communication phase. Unlike the preprocessing step, each client considers $\Omega_k$ during training, which does not exceed the budget constraint. Therefore, GGC becomes more efficient during training by requiring less communication than BGGC and adhering to the budget constraint.
\end{remark}

\vspace{-0.2cm}
\subsection{Properties of the proposed optimization framework}
\vspace{-0.1cm}
\begin{proposition}
\label{proposition1}
The collaboration graph resulting from solving \cref{eq:perso_collaborators_initial} yields superior or same results compared to any collaboration graph $\mathcal{C} \in \{\mathcal{C}_i \in 2^{[N] \setminus {i}}; |\mathcal{C}_i| \leq B_c\}_{i=1}^N$ that imposes additional restrictions on the collaboration structure, where $N$ is the total number of clients.
\end{proposition}
\vspace{-0.1cm}
The proof of \cref{proposition1} is in \cref{app:proof_proposition}.
\begin{remark}
    It follows from \cref{proposition1} that our targeted collaboration graph outperforms pure local training (no one collaborates with anyone), any random graph within $\mathcal{C}$, and any graph with further imposed symmetry.
\end{remark}

\begin{remark}
Similar to \cite{shalev2009stochastic, feldman2019high}, we assume that the validation loss is a good approximation of $F_k(\mathbf{w}_k)$. Hence, \cref{eq:greedy_eq_t} is a good proxy for \cref{eq:perso_collaborators_initial}. Additionally, we use \cref{alg:GGC}, which offers approximation guarantees even with noisy reward functions, as demonstrated in \cite[Corollary 2]{fourati2023randomized}. Moreover, it returns a set no worse than the empty set with probability 1. Therefore, DPFL using this algorithm for graph generation is expected to yield a good solution to \cref{eq:perso_collaborators_initial}. 
\end{remark}

\vspace{-0.2cm}
\section{Experiments}
\vspace{-0.2cm}
We present the key experimental settings and results. We provide additional details in \cref{app:detailed_exp}. 
\vspace{-0.2cm}
\label{sec:Experiments}
\subsection{Setup}
\vspace{-0.2cm}
\label{sec:setup}
We adopt commonly used datasets from existing literature on personalized FL~\cite{ditto, fedrep, marfoq2022personalized}. We conduct experiments with CIFAR10 \cite{krizhevsky2009learning}, Federated Extended MNIST (FEMNIST) \cite{caldas2018leaf}, and CINIC10 \cite{darlow2018cinic}. With CIFAR10 and CINIC10 we consider two scenarios for partitioning the datasets according to different heterogeneous distributions. The first scenario, denoted as $Patho(3)$, involves a pathological distribution split \cite{zhang2023fedala, mcmahan2017communication, colin2016gossip}, wherein each client exclusively receives data from three specified categories. The second scenario utilizes the Dirichlet distribution \cite{yurochkin2019bayesian, wang2020federated}, where a distribution vector $\mathbf{q}_{c}$ is drawn from $Dir_k(\alpha)$ for each category $c$. Subsequently, the proportion $\mathbf{q}_{c,i}$ of data samples from category $c$ is allocated to client $i$. For FEMNIST, we consider the natural non-IID split provided in the Leaf framework \cite{caldas2018leaf} where each writer corresponds to a client and we add more degrees of heterogeneity by having each client missing some classes.
For CIFAR10 and FEMNIST, we conduct 100 rounds of training, while for CINIC10 we run for 50 rounds.
All experiments are executed across three different random seeds; the results report the average local test accuracy along with the standard deviation. We preserve a best model per client based on the validation dataset, and the reported results are obtained by performing inference on the testing set using such best-validation models. 
Other experimental details, including hyperparameters like the numbers of initialization and training epochs ($\tau_{\text{init}}$ and $\tau_{\text{train}}$) are in the appendix.

\vspace{-0.1cm}
\subsection{Quality of personalization results}
\vspace{-0.1cm}

\begin{table*}[t]
\centering
\small
\resizebox{\textwidth}{!}{
\begin{tabular}[t]{|l|rr|rr|r|}
\hline & \multicolumn{2}{|c|}{ CIFAR10 } & \multicolumn{2}{c|}{ CINIC10 } & FEMNIST  \\
\hline  & $Dir(0.1)$ & $Patho(3) $  & $Dir(0.1)$ & $Patho(3)$ & Natural Split\\
\hline Local Only & 80.38 $\pm$ 1.62 & 78.32 $\pm$ 0.49 & 78.59 $\pm$ 1.00 & 77.53 $\pm$ 0.47 & 87.27 $\pm$ 0.27\\
\hline FedAvg \cite{mcmahan2017communication} & 47.22 $\pm$ 1.20 & 50.93 $\pm$ 1.97  & 38.48 $\pm$ 0.37 & 39.27 $\pm$ 1.29 & 87.41 $\pm$ 0.72\\
FedAvg+FT & 82.84 $\pm$ 0.88 & 81.89 $\pm$ 0.92 & 80.31 $\pm$ 0.33 & 79.18 $\pm$ 0.46
 & 91.56 $\pm$ 0.29\\
FedProx \cite{lian2018asynchronous} & 48.83 $\pm$ 1.88 & 51.14 $\pm$ 0.79 & 38.16 $\pm$ 0.82 & 38.53 $\pm$ 0.54 &88.14 $\pm$ 0.39\\
FedProx+FT \cite{lian2018asynchronous} & 78.86 $\pm$ 0.51 & 73.61 $\pm$ 1.29 & 76.59 $\pm$ 1.19 & 73.04 $\pm$ 1.08 & 93.58 $\pm$ 0.29\\
\hline APFL \cite{smith2017federated} & 82.56 $\pm$ 0.38 & 80.96 $\pm$ 0.71 & 80.02 $\pm$ 1.07 & 78.21 $\pm$ 0.39 & 90.80 $\pm$ 0.37\\
PerFedAvg \cite{fallah2020personalized} & 82.38 $\pm$ 1.43 & 81.17 $\pm$ 0.52 & 78.88 $\pm$ 1.25 & 78.90 $\pm$ 0.37 & 91.69 $\pm$ 0.40\\
Ditto \cite{ditto} & 83.10 $\pm$ 0.70 & 81.19 $\pm$ 0.63 & 80.33 $\pm$ 1.27 & 79.52 $\pm$ 0.31 & 90.83 $\pm$ 0.46\\
FedRep \cite{fedrep} & 80.81 $\pm$ 1.09 & 78.61 $\pm$ 1.67  & 79.56 $\pm$ 1.12 & 78.46 $\pm$ 0.69  & 90.04 $\pm$ 0.28\\
kNN-Per \cite{marfoq2022personalized} & 82.45 $\pm$ 0.69 & 81.04 $\pm$ 0.40  & 80.42 $\pm$ 1.07 & 79.54 $\pm$ 0.32  & 91.29 $\pm$ 0.40\\
pFedGraph \cite{ye2023personalized} & 80.48 $\pm$ 0.75 & 78.34 $\pm$ 1.04 & 78.73 $\pm$ 1.04 & 77.58 $\pm$ 0.16 & 87.70 $\pm$ 0.13\\
\hline \textbf{DPFL} ($B_c= \inf$) & $\mathbf{84.39}$ $\pm$ 0.43 & $\mathbf{83.04}$ $\pm$ 0.98 & $\mathbf{81.49} \pm$ 1.32 & $\mathbf{80.32} \pm$ 0.51 & $\mathbf{94.25}$ $\pm$ 0.18\\
\textbf{DPFL} ($B_c=0.2N$) & $\mathbf{84.01}$ $\pm$ 0.44 & $\mathbf{82.83}$ $\pm$ 0.91  &  $\mathbf{81.24} \pm$ 1.30  & $\mathbf{80.51} \pm$ 0.39 & $\mathbf{94.31}$ $\pm$ 0.06 \\
 \textbf{DPFL} ($B_c=0.1N$) & $\mathbf{83.86}$ $\pm$ 0.46 & $\mathbf{82.52}$ $\pm$ 0.78& $\mathbf{81.32} \pm$ 1.20 & $\mathbf{80.24} \pm$ 0.35  & $\mathbf{94.09}$ $\pm$ 0.14\\
 \textbf{DPFL} ($B_c=0.05N$) & $\mathbf{82.91}$ $\pm$ 0.81 & $\mathbf{82.17}$ $\pm$ 0.54 & $\mathbf{80.91} \pm$ 1.31 & $\mathbf{80.11} \pm$ 0.30 & $\mathbf{93.82}$ $\pm$ 0.17\\
\hline
\end{tabular}}
\caption{Comparison of accuracy across benchmarks is illustrated as $acc \pm std_r$, where acc represents the average test accuracies of all clients' models, and $std_r$ denotes the standard deviation over three repetitions with different random seeds.}
\vspace{-2.em}
\label{tab:Accu_image}
\end{table*}

\Cref{tab:Accu_image} presents the average test accuracies of DPFL under four budget constraints, compared to other personalized FL and baseline methods. For DPFL, we vary the budget constraint with 20\% ($B_c = 0.2N$), 10\% ($B_c = 0.1N$), and 5\% ($B_c = 0.05N$) of the total number of clients, respectively; we also consider the case without a constraint ($B_c = \text{inf}$).

The results demonstrate that our method significantly outperforms other personalized methods regarding average test accuracy. We achieve better results than local training by 3 and 5 percentage points (p.p), outperform FedAvg by 37 and 33 p.p, and surpass other personalized methods by approximately 1 to 6 and 2 to 10 p.p on CIFAR10 using $Dir(0.1)$ and $Patho(3)$, respectively.

Moreover, on FEMNIST, we achieve better results than local training by 7 p.p, outperform FedAvg by 7 p.p, and surpass other personalized methods by approximately 2.8 to 6.5 p.p. Finally, on CINIC10, our method demonstrates superior performance, surpassing local training by 3 p.p in both $Dir(0.1)$ and $Patho(3)$. Additionally, DPFL outperforms FedAvg by a notable margin of 43 and 41 p.p, while surpassing other personalized methods by approximately 1 to 5 and 2 to 7 p.p using $Dir(0.1)$ and $Patho(3)$, respectively.

\begin{wrapfigure}[14]{r}{0.38\textwidth}
\vspace{-0.2cm}
\begin{tikzpicture}
  \node (img)  {\includegraphics[scale=0.28]{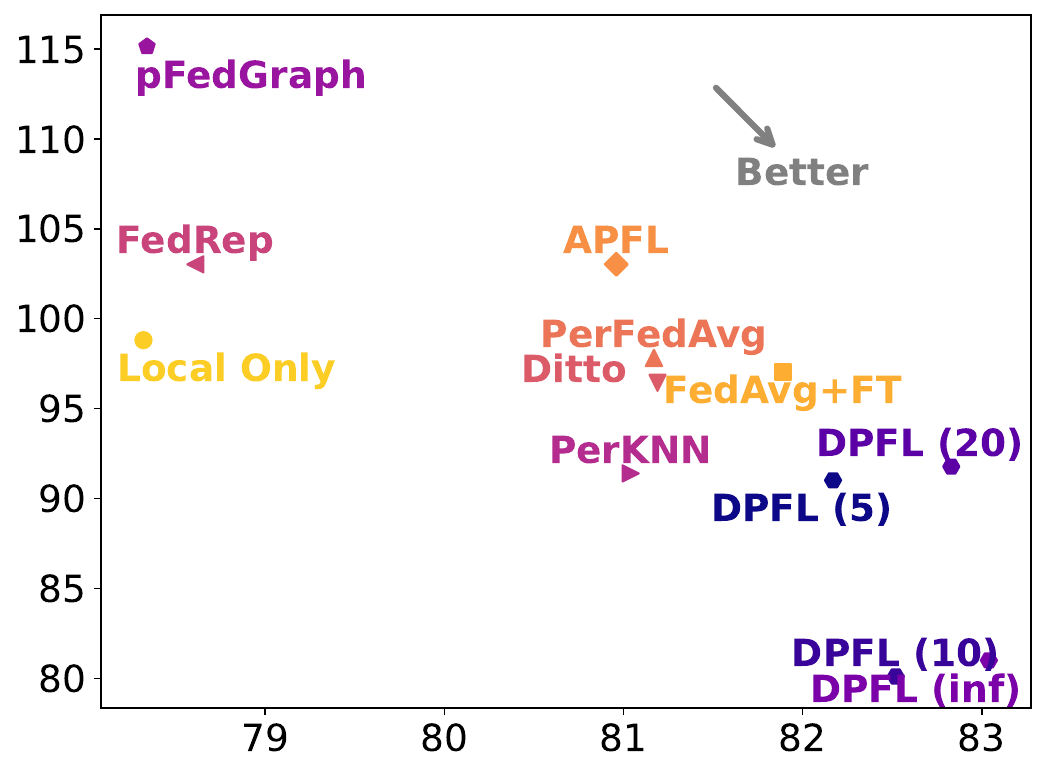}};
  \node[node distance=0cm, rotate=90, anchor=center,yshift=2.6cm] {\tiny Variance};
  \node[node distance=0cm, rotate=0cm, anchor=center,yshift=-2.0cm] {\tiny Test Accuracy};
\end{tikzpicture}
\caption{Variance between local models using Patho(3) data splits. 
}
\label{fig:CIFAR10_variance}
\end{wrapfigure}
\smartparagraph{Variance between local models.}
As our objective is personalization, apart from the overall improvement in average accuracy among clients, it is crucial to assess whether improvements are distributed across most models rather than confined to just a few clients. To evaluate this, we analyze the variance between local models, where lower variance signifies greater parity in their performance.
\Cref{fig:CIFAR10_variance} shows results for CIFAR10 with a $Patho(3)$ distribution. The x-axis represents average test accuracy, and the y-axis represents variance between clients' models. DPFL is consistently in the right-bottom corner, showing higher overall accuracy and lower variance than other methods. Additionally, \cref{appendix:Variance} confirms these findings for FEMNIST (\cref{fig:femnist_variance}), CINIC10 (\cref{fig:CINIC10_variance}), and CIFAR10 with a $Dir(0.1)$ distribution (\cref{fig:CIFAR10_variance_appendix}).

\vspace{-0.2cm}
\subsection{Visualization of collaboration graph}
\vspace{-0.2cm}
We analyze the initial collaboration graph and its evolution after 50 and 99 rounds.
For clarity, we illustrate two cases with budget $B_c = 10$ and $B_c = 5$ (other budget constraints are in \cref{appendix:Col_graph}).
\cref{fig:collaboration_graph_20,fig:collaboration_graph_5} depict the collaboration graph (plotted as the adjacency matrix) in two states: (left) constructed as a preprocessing step (line 5 of \cref{alg:DPFL}), and (right) at round 99.
The diagonal indicates that every client always "collaborates" with itself.
To illustrate the graph evolution, the right plots display collaborative links in two colors: in red are the clients selected for collaboration in that round; in blue are the clients identified during the preprocessing step but are currently not chosen.
The union of red and blue clients corresponds to the initial collaboration graph.

The figures highlight that the initial collaboration graph is denser compared to the actually used clients for aggregation in round 99. This is expected since the graph is constructed as a preprocessing step, and at this stage, model weights have not yet converged. Therefore, broader collaboration can be beneficial. However, as training progresses, it is natural to expect each client to benefit primarily from collaborating with clients with similar data distributions, thus leading to a sparser collaboration graph. Another contributing factor is that, in the preprocessing step, the decision to select a specific client for collaboration is made from a pool of 100 clients, making it more challenging than in later rounds where the decision is drawn from a smaller pool denoted as $\Omega_{k}$ for client $k$.

\begin{figure}[t]
  \centering
  \begin{subfigure}[t]{0.49\textwidth}    \includegraphics[scale=0.32]{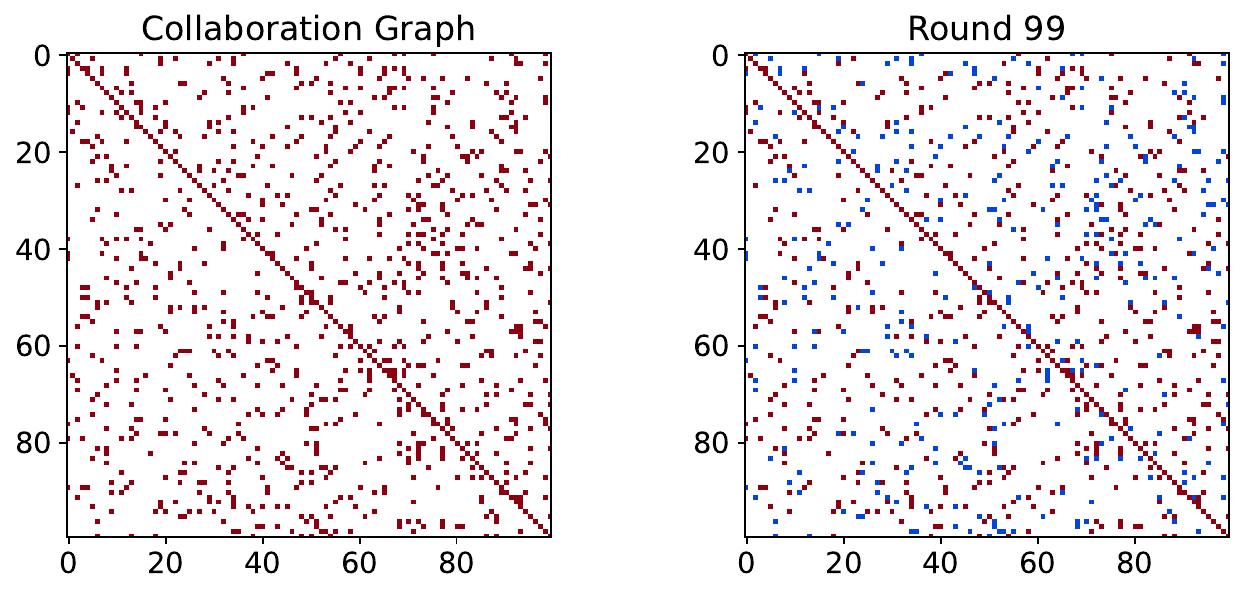}
    \caption{\small $B_c=10$}
\label{fig:collaboration_graph_20}
  \end{subfigure}
  \begin{subfigure}[t]{0.49\textwidth}
\includegraphics[scale=0.32]{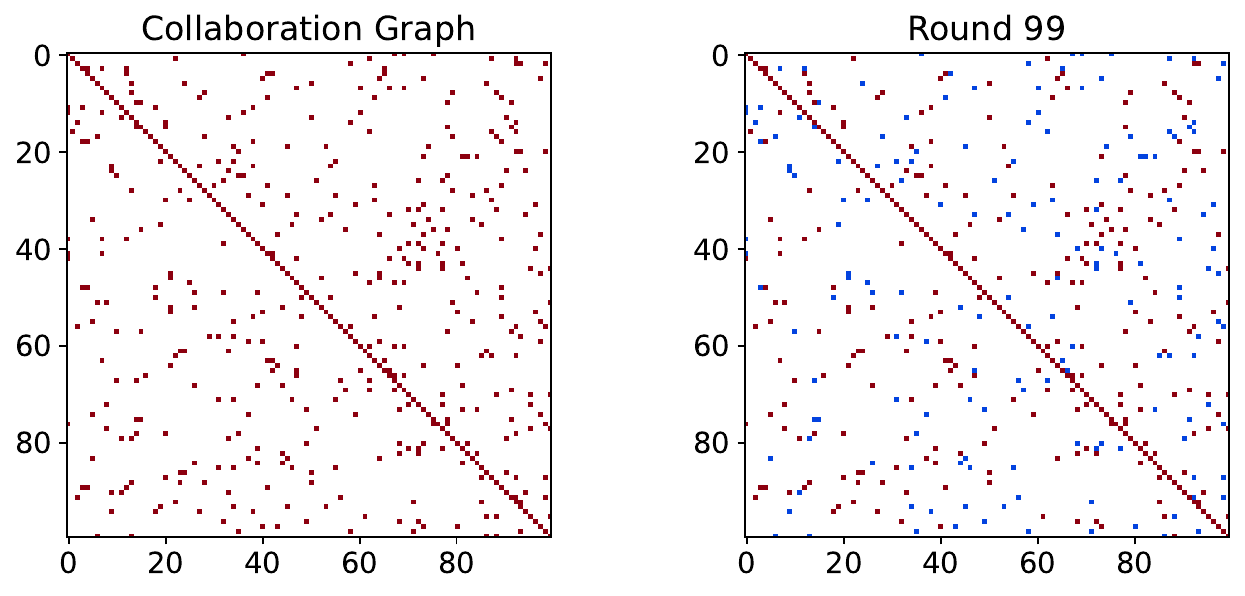}
    \caption{\small $B_c=5$}
\label{fig:collaboration_graph_5}
  \end{subfigure}
  \caption{\small Collaboration graph using CIFAR10 with 100 clients.}
\end{figure}

Finally, we analyze the graph sparsity. With $B_c = 10$,  initial sparsity stands at 80\% and it increases to 88\% at round 99. With $B_c = 5$, the initial sparsity is 95\% and is 96\% at round 99. We also measure the symmetry of the collaboration graph across different budgets (details in \cref{app:Asymmetry}); we observe around 80-88\% symmetry, which decreases as the budget constraint lowers.

\newpage

\begin{wrapfigure}[6]{r}{0.35\textwidth}
\vspace{-1.8cm}
\begin{center}
\begin{tikzpicture}
\node (img) {\includegraphics[scale=0.33]{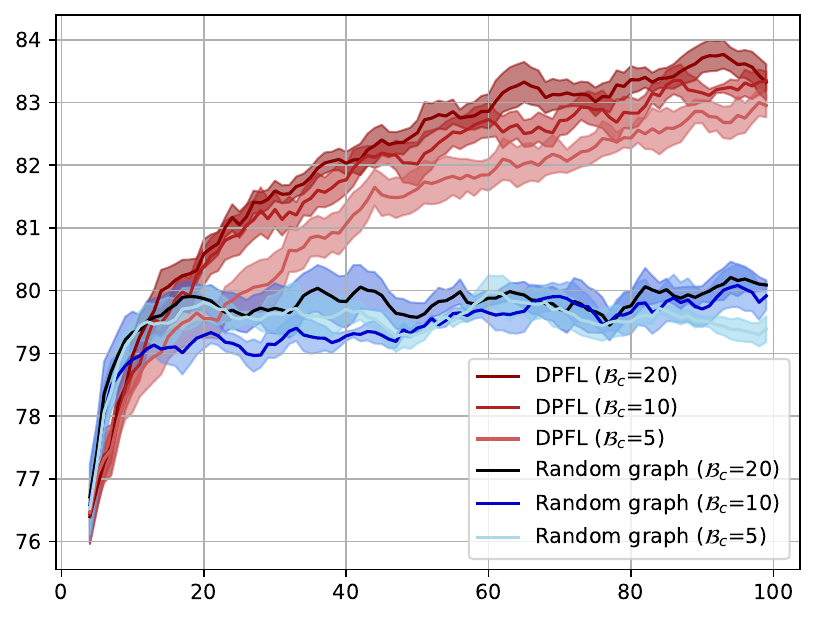}};
\node[rotate=90, anchor=center,yshift=2.4cm] {\tiny Test Accuracy};
\node[rotate=0cm, anchor=center,yshift=1.8cm] {\tiny Round};
\end{tikzpicture}
\caption{\small DPFL vs. random graph.
}
\label{fig:DPFL_vs_random}
\end{center}
\end{wrapfigure}
\subsection{Comparison to a randomly-generated graph}
To illustrate the relative importance of GGC, we compare DPFL to a version of our method that replaces GGC with a randomly-constructed collaboration graph. \cref{fig:DPFL_vs_random} shows that DPFL outperforms the random graph by $\approx$4 p.p, in three budget constraints.
This experiment uses CIFAR10 with 100 clients and $Dir(0.1)$.

\subsection{Behavior of DPFL under data flip attack}
To analyze how DPFL behaves in the presence of distinct groups of clients, we select 40 clients (malicious) out of 100 and flip their labels using the same permutation; the remaining 60 clients (benign) use the true labels (according to CIFAR10).
It's important to note that our objective is delineate the behavior of DPFL, acknowledging that while it may exhibit robustness characteristics, the study of robustness falls outside the scope of this paper.

\begin{figure}[t!]
  \centering
  \begin{subfigure}[t]{0.49\textwidth}
    \centering
    \includegraphics[scale=0.33]{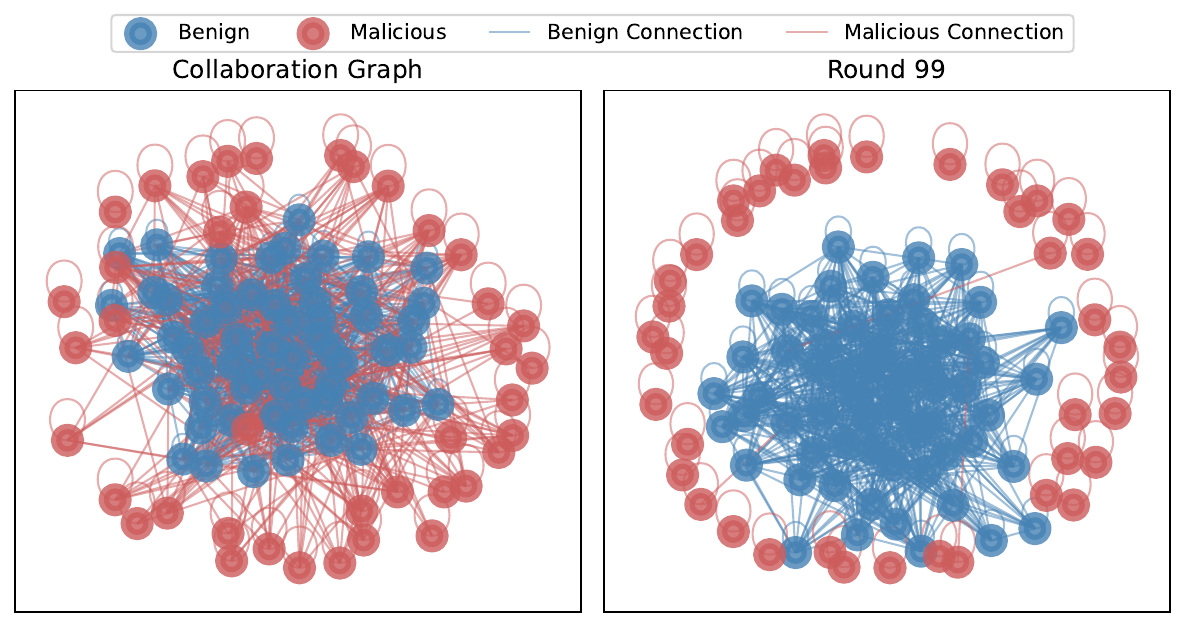}
    \caption{\small Malicious clients don't execute GGC.}
\label{fig:malicious_benign_without_greedy}
  \end{subfigure}
  \begin{subfigure}[t]{0.49\textwidth}
  \centering
    \includegraphics[scale=0.33]{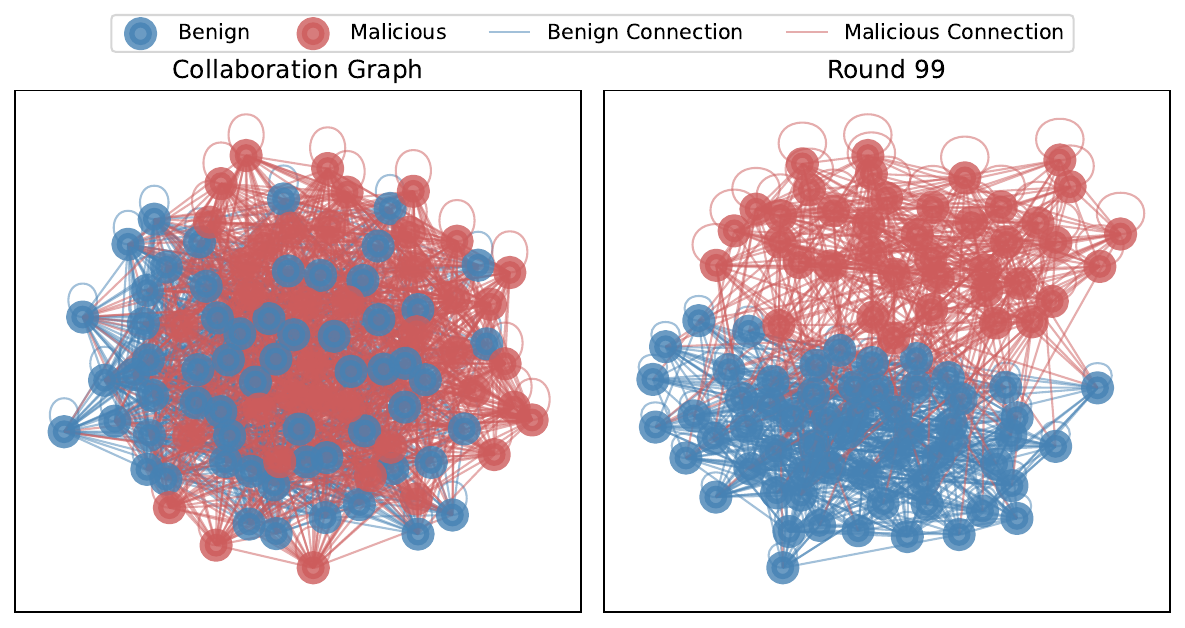}
    \caption{\small Malicious clients execute GGC.}
\label{fig:malicious_benign_with_greedy}
  \end{subfigure}
  \caption{ \small Collaboration graph when 40\% of clients have flipped labels (malicious), while 60\% have original labels (benign). For both scenarios, we show the initial collaboration graph (left) and its evolution after 99 rounds (right). Malicious clients appear in red; benign ones are in blue.}
\vspace{-1em}
\end{figure}

We then conduct two experiments: the first in which malicious clients do not execute GGC (i.e., they only train locally), \cref{fig:malicious_benign_without_greedy}; the second in which they run GGC, \cref{fig:malicious_benign_with_greedy}. In the first case, the collaboration graph initially contains numerous edges involving malicious clients. This is expected and is due to the inherent randomness in the weights during the preprocessing step, which makes it challenging to identify collaborators. However, as the rounds progress, we observe that benign clients increasingly avoid selecting malicious ones until, ultimately, they cease choosing them altogether. \cref{fig:full_graph_attack_without_greedy} depicts this evolution every 10 rounds.

In the second case, since malicious clients execute GGC, they initially collaborate with benign clients. Thus, their models become regularized towards the benign ones. Despite this behavior, we observe that as the rounds progress, clients become almost segregated into two groups (red and black), with very few links between them serving as a form of regularization (full evolution in \cref{fig:full_graph_attack_with_greedy}). On average, benign clients have less than 10\% connections with malicious ones (see \cref{fig:full_graph_attack_with_greedy} and \cref{fig:histo}).

\vspace{-0.2cm}
\subsection{Ablation studies}
\textbf{Sensitivity to $\tau_{\text{init}}$.}
\cref{tab:sensitivity_analysis_1} reports the results of using CIFAR10 with 100 clients and $Patho(3)$ split and varying the number of local epochs $\tau_{\text{init}}$. The results show that the accuracies across different numbers of local epochs are comparable. Performance on average is slightly better with $\tau_{\text{init}}=10$, but even the case with $\tau_{\text{init}}=1$ yields good results.

\begin{table}[t!]
\centering
\resizebox{0.7\linewidth}{!}{
\begin{tabular}[h]{|r|r|r|r|r|}
\hline
\diagbox[width=5em]{$\tau_{\text{init}}$}{$B_c$} & \thead{inf} & \thead{20} & \thead{10} & \thead{5} \\
\hline
1 &$81.86 \pm 1.00$ & $82.33 \pm 0.86$ & $82.53 \pm 0.72$ & $\mathbf{82.30} \pm 0.24$ \\
5 &$83.03 \pm 0.73$ & $82.53 \pm 1.36$ & $\mathbf{82.58} \pm 0.97$ & $80.90 \pm 0.96$ \\
10 &$\mathbf{83.04} \pm 0.98$ & $\mathbf{82.83} \pm 0.91$ & $82.52 \pm 0.78$ & $82.17 \pm 0.54$ \\
\hline
\end{tabular}}
\captionsetup{skip=15pt}  
\caption{Effect of local epochs performed in the preprocessing step on the convergence of DPFL.
\label{tab:sensitivity_analysis_1}}
\vspace{-2em}
\end{table}

\begin{wrapfigure}[9]{r}{0.38\textwidth}
\vspace{-0.4cm}
\resizebox{\linewidth}{!}{
\begin{tabular}{|r|r|r|r|r|}
\hline
\diagbox[width=5em]{$P$}{$B_c$} & \thead{inf} & \thead{20} & \thead{10} & \thead{5} \\
\hline
1  & 84.20 &83.67 & 84.20 & 84.33 \\ 
5  & 83.83 &84.01 & 83.31 & 83.73 \\ 
10 & 83.24 &84.17 & 83.31 & 83.03\\ 
20 & 82.34 &82.59 & 82.08 & 82.73 \\ 
\hline
\end{tabular}
}
\captionof{table}{Effect of the periodicity of invoking GGC on the convergence of DPFL.}
\label{tab:sensitivity_analysis_2}
\end{wrapfigure}

\smartparagraph{Periodicity of refreshing $\mathcal{C}_{k}$.}
To improve efficiency, we consider invoking GGC at line 10 in \cref{alg:DPFL} periodically every $P$ rounds.
\cref{tab:sensitivity_analysis_2} reports the effect of varying $P$ while training on CIFAR10 using 100 clients and $Dir(0.1)$.

We observe that DPFL maintains good performance across various periodicities, with a slight decrease as $P$ increases. This demonstrates the robustness of our method and the potential for improved efficiency by performing it periodically. 

\vspace{-0.1cm}
\section{Related Work}
\vspace{-0.1cm}
\label{related_work}
Personalized FL approaches address data heterogeneity through various techniques. Some methods train separate global and personalized models \cite{ditto, pfedme} and regularize them using an $l_2$ term. Others leverage multi-task learning \cite{hanzely2020federated, khodak2019adaptive, jiang2019improving, mansour2020three, pmlr-v151-gasanov22a, hanzely2020lower, marfoq2022personalized}, which are based on interpolation between local and global models. Meta-learning-based methods, such as \cite{perfedavg}, propose to find a common initial model that new clients can easily adapt. Splitting model layers \cite{fedrep, liang2020think} or using hypernetworks for personalization are also explored \cite{shamsian2021personalized, chen2021bridging}. Compared to these methods, we eliminate potential communication overhead by not requiring a global model. Furthermore, we achieve a fine-grained level of personalization by dynamically selecting the most relevant collaborators for each client. 

Furthermore, unlike cosine similarity-based methods \cite{ye2023personalized, li2022learning, li2022towards}, which relies on pairwise collaborator selection, as discussed in \cref{sec:greedy_alg}, overlooking the significance and synergy of the entire group. Instead, we propose a novel utility function (\cref{eq:greedy_eq_t}) that considers the combined benefit of collaborators. Furthermore, unlike our method which considers a budget constraint, in \cite{ye2023personalized}, all the clients will be considered with a weighted average, which limits its scalability for real-world constrained scenarios. Finally, personalized decentralized approaches like \cite{vanhaesebrouck2017decentralized, koppel2017proximity, even2022sample} address heterogeneity but assume a given collaboration graph and are often limited to linear models or pre-trained model combinations. They also rely on parameter tuning for graph sparsity, unlike our exact budget.

\vspace{-0.1cm}
\section{Limitations}
\vspace{-0.1cm}
\label{sec:limitations}
Our approach considers uniform resource and communication budgets for all clients $B_c$, while in practice, clients may have varying resource capabilities $B_c^i$, for client $i$. Furthermore, this work assumes the possibility of choosing any available client for communication. However, some clients may not be within communicable distance of others, preventing establishing a connection. These limitations can be addressed by considering personalized budgets for each client and restricting the potential set of collaborators to only those within communicable distance.
\vspace{-0.1cm}
\section{Conclusion}
\vspace{-0.1cm}
We address the challenge of constructing a collaboration graph in decentralized learning while considering data heterogeneity and adhering to communication and resource constraints. To achieve this, we propose a bi-level optimization problem and employ a combinatorial solution using a greedy algorithm to efficiently identify the collaboration graph. We compared our solution to state-of-the-art methods and baselines on different datasets and demonstrated its superior performance.

\bibliography{main}
\bibliographystyle{plain}

\newpage
\appendix
\onecolumn

\section{Motivation for Our Combinatorial Proposition}

\label{appendix:motivation}
In this section, we will show how our utility function based on group synergy by considering the combinatorial effect of clienwts is more interesting than pairwise comparisons.

We conducted an experiment utilizing the CIFAR10 dataset. We allocated three clients as follows: Client 1 has access to four classes (0, 4, 6, 8) with 50, 50, 50, and 50 data points per class, respectively. Client 2 is assigned four classes (0, 6, 1, 3) with 300, 300, 200, and 200 data points per class, respectively. Lastly, client 3 has access to four classes (4, 8, 5, 7) with 300, 300, 200, and 200 data points per class, respectively. In other words, clients 2 and 3 have two classes in common each with client 1.
We then train three models using FedAvg-fine-tuned-like training (E local steps and aggregation): one where clients 1 and 2 collaborate, one where clients 1 and 3 collaborate, and one where all three clients collaborate.
Remarkably, our findings indicate that collaboration between clients 1 and 2, or 1 and 3 alone, leads to a decrease in performance by approximately 11\% in both cases when compared to client 1 training a model independently on its own private dataset alone. 

However, the collaborative effort involving clients 1, 2, and 3 collectively enhances accuracy by approximately 6\%. 

The effectiveness of GGC in building the collaboration graph could also be shown in the case of two clients each of them contributed positively to client 1, however the collaboration of 1,2,3 leads to negative gain. 

Moreover, we would like to clarify that these two cases are the extreme ones where the algorithms that compare pairwise performance fail to address, however sometimes adding client 1 alone is good and adding client 2 alone is good, but the marginal gain of adding them together is significantly higher, and this could play an important role when we have communication constraints where we are restricted to collaborate only with a subset of clients and choosing them wisely is important.

We believe that scenarios, where the synergy between a group of clients outweighs the pairwise comparison of collaboration effectiveness, are common in real-world settings, particularly when dealing with highly heterogeneous data. In such cases, a specific client might adversely affect the training of another client. However, when collaborating with a "complementary client", the combined effort can significantly enhance performance.

\section{Greedy Graph Construction (GGC)}
\label{app:GGC}
\begin{algorithm}[h]
\caption{Greedy Graph Construction (GGC)}\label{alg:GGC}
\begin{algorithmic}[1]
\REQUIRE client $k$, clients $\mathcal{S}$
        \STATE $\mathcal{X} \leftarrow \{k\}, \mathcal{Y}\leftarrow \mathcal{S}\cup \{k\},$  \\
        \FOR{client $j \in \text{Shuffle}(\mathcal{S})$}
        \STATE  $\mathcal{R}(\mathcal{X}) \leftarrow - F_k^{\mathcal{V}}\left(\frac{1}{\sum_{i \in \mathcal{X}} p_i} \sum_{i \in \mathcal{X}} p_i \mathbf{w}_i\right)$
        \STATE  $\mathcal{R}(\mathcal{X}\cup \left\{j\right\}) \leftarrow - F_k^{\mathcal{V}}\left(\frac{1}{\sum_{i \in \mathcal{X} \cup \{j\}} p_i} \sum_{i \in \mathcal{X} \cup \{j\}} p_i \mathbf{w}_i\right)$
    \STATE  $\mathcal{R}(\mathcal{Y}) \leftarrow - F_k^{\mathcal{V}}\left(\frac{1}{\sum_{i \in \mathcal{Y}} p_i} \sum_{i \in \mathcal{Y}} p_i \mathbf{w}_i\right)$
        \STATE  $\mathcal{R}(\mathcal{Y}\setminus \left\{j\right\}) \leftarrow - F_k^{\mathcal{V}}\left(\frac{1}{\sum_{i \in \mathcal{Y} \setminus \{k\}} p_i} \sum_{i \in \mathcal{Y} \setminus \{j\}} p_i \mathbf{w}_i\right)$
        
            \STATE $a \leftarrow \max(\mathcal{R}(\mathcal{X} \cup \left\{j\right\}) - \mathcal{R}(\mathcal{X}), 0)$
            \STATE $b \leftarrow \max(\mathcal{R}(\mathcal{Y}\setminus \left\{j\right\}) - \mathcal{R}(\mathcal{Y}), 0)$
            \STATE \textbf{with probability} $p = \frac{a}{a + b}$ \textbf{do} 
            \STATE $\quad$ $\mathcal{X} \leftarrow \mathcal{X} \cup \left\{j\right\}$ and $\mathcal{Y} \leftarrow \mathcal{Y}$
            \STATE \textbf{else}
            \STATE $\quad$ $\mathcal{Y} \leftarrow \mathcal{Y} \setminus \left\{j\right\}$ and $\mathcal{X}\leftarrow \mathcal{X}$
        \ENDFOR
        \IF{$|\mathcal{X}| = B_c$}
        \STATE  \textbf{break}
    \ENDIF
    \STATE \textbf{Return} $\mathcal{X}$
\end{algorithmic}
\end{algorithm}

\textbf{Additional explanation}: 
We would like to note that as our objective function is not monotone, meaning that adding won't always be the best choice, we necessitate the computation of both marginal gains of adding and removing a client. To this end, four cases will appear, three of them will happen or do not happen with probability 1. 

\begin{itemize}
    \item If the marginal gain of adding is positive (a is positive) and the marginal gain of removing is negative (b is zero) then p = 1. 

    \item If the marginal gain of adding is negative (a is zero) and the marginal gain of removing is positive (b is positive) then p = 0 (does not happen with probability 1).

    \item  If the marginal gain of adding is negative (a is zero) and the marginal gain of removing is negative (b is zero) then p = 1.

    \item If the marginal gain of adding is positive (a is positive) and the marginal gain of removing is positive (b is positive) only in this case we flip a biased coin that decides probability p based on the values of a and b for either adding or removing that client. In our experiments, we noticed that this case happens less than 1\% of the time.

\end{itemize}

Furthermore, it has been shown in \cite{buchbinder2015tight} that if you remove the randomness in the agnostic case i.e., when the marginal gains of adding and removing are both positive (a and b are positive) then the algorithm guarantees decreases from achieving $\frac{1}{2}$ of the optimal solution to the $\frac{1}{3}$ of the optimal solution.

\section{Efficient Reward Computation}
\label{appendix:efficient_calls}

Looking at \cref{alg:GGC}, we notice that it requires reward computations to the function $\mathcal{R}(S)$, which is defined in our objective according to \cref{eq:reward}. This reward computation requires access to $\mathbf{w}_i$ in $\mathcal{S}$, where $|\mathcal{S}|$ may exceed our budget constraint $B_c$ (see \cref{eq:preprocessing}). Therefore, to ensure that we do not violate our constraints during the preprocessing step outlined in \cref{alg:DPFL}, we propose amortizing the communication complexity needed to evaluate the objective function. This involves dividing the necessary downloads into $O(\frac{n}{B_c})$ steps. Each step entails only $O(B_c)$ computation, storage, and communication complexity, as opposed to a single step with $O(n)$ complexity, which breaches the budget constraint. To elaborate, during each communication step, a client $k$ receives at most $B_c$ model updates and monitors the necessary averages for algorithmic decision-making, without retaining all individual models.

To execute this, the process commences by preparing the call for variable $b$, requiring the average of all models in $\mathcal{Y}$, beginning with a full set of clients. We envisage $\lceil \frac{n}{B_c} \rceil$ communication steps (line 2), where in each step, $B_c$ clients transmit their model updates (line 3). We compute the average, as depicted in lines 4-6 of \cref{alg:BGGC}. In line 5, $\mathbf{w}^{Y}$ denotes storing the weighted-sum of received models in batches. Upon completing these steps, $\mathbf{w}^{Y}$ is computed. The second phase of communications initiates. Client $k$ begins receiving batches of models without replacement from $B_c$ clients, storing their indices in $S_b$. Similar to GGC, client $k$ computes the marginal gains of adding and removing a client from $\mathcal{S}_b$ by computing $a$ and $b$, respectively.  A decision is then made based on probability $p$ in line 16. A crucial step is to keep track of the values of $\mathbf{w}^{X}$ and $\mathbf{w}^{Y}$ and update them as in lines 17 and 19, for future use even for the next received batch of models $B_c$.

It's important to note that this expanded window of communications is solely required for the preprocessing step; for graph verification at line 10 of Algorithm \ref{alg:DPFL}, the algorithm takes as input $|\Omega_k| \leq B_c$.

\begin{algorithm}[h]
\caption{Batched Greedy Graph Construction (BGGC)}\label{alg:BGGC}
\begin{algorithmic}[1]
\REQUIRE client $k$, clients $\mathcal{S}$, budget $B_c$. 
    \STATE $\mathbf{w}^{Y} \leftarrow p_k \mathbf{w}_k$ 
\FOR{$s$ in range($\lceil \frac{n}{B_c} \rceil$)}
        \STATE \text{Client $k$ receives a batch $\mathcal{B}$ of at most $B_c$ models without replacement}
        \FOR{model $\mathbf{w}_s \in \mathcal{B}$}
            \STATE $\mathbf{w}^{Y}\leftarrow \mathbf{w}^{Y} + p_s * \mathbf{w}_s$
    \ENDFOR
    \ENDFOR
        \STATE $\mathcal{X} \leftarrow \{k\}, \mathcal{Y}\leftarrow \mathcal{S}\cup \{k\},$ $\mathbf{w}^{X}\leftarrow p_k \mathbf{w}^k, \mathbf{w}^{Y}\leftarrow \mathbf{w}^{Y},$ \\

        \STATE $\mathcal{S} \leftarrow$ Shuffle($\mathcal{S}$)\\
        \FOR{$s$ in range($\lceil \frac{n}{B_c} \rceil$)}
        
        \STATE \text{Client $k$ receives a batch $\mathcal{B}$ of at most $B_c$ models in order from $\mathcal{S}$} \\
        \text{without replacement and indices in $\mathcal{S}_b$}
        \FOR{j in $\mathcal{S}_b$} 
        \STATE $\mathcal{R}(\mathcal{X}) \leftarrow -F_k^{\mathcal{V}}\left(\frac{\mathbf{w}^{X}}{\sum_{i \in \mathcal{X}} p_i}\right)$\\
        \STATE$ \mathcal{R}(\mathcal{X} \cup \left\{j\right\}) \leftarrow -F_k^{\mathcal{V}}\left(\frac{\mathbf{w}^{X} + p_j \mathbf{w}^{j}}{(p_j + \sum_{i \in \mathcal{X} } p_i)} \right)$\\
        \STATE $\mathcal{R}(\mathcal{Y}\setminus \left\{j\right\}) \leftarrow -F_k^{\mathcal{V}}\left(\frac{\mathbf{w}^{Y} - p_j \mathbf{w}^{j}}{(- p_j + \sum_{i \in \mathcal{Y} } p_i)} \right)$\\
            \STATE $\mathcal{R}(\mathcal{Y}) \leftarrow -F_k^{\mathcal{V}}\left(\frac{\mathbf{w}^{Y}}{\sum_{i \in \mathcal{Y}} p_i} \right)$\\
            \STATE $a \leftarrow \max(\mathcal{R}(\mathcal{X} \cup \left\{j\right\}) - \mathcal{R}(\mathcal{X}), 0)$ $\quad$ \COMMENT{marginal gain of adding a client}\\
            \STATE $b \leftarrow \max(\mathcal{R}(\mathcal{Y}\setminus \left\{j\right\}) - \mathcal{R}(\mathcal{Y}), 0)$ $\quad$ \COMMENT{marginal gain of removing a client}
            \STATE \textbf{with probability} $p = \frac{a}{a + b}$ \textbf{do} 
            \STATE $\quad$ $\mathcal{X} \leftarrow \mathcal{X} \cup \left\{j\right\}$ \text{;} $\quad$ $\mathcal{Y} \leftarrow \mathcal{Y}$ ; $\quad$ $\mathbf{w}^{X} \leftarrow \mathbf{w}^{X} + p_j \mathbf{w}^{j}$ 
            \STATE \textbf{else}
            \STATE $\quad$ $\mathcal{Y} \leftarrow \mathcal{Y} \setminus \left\{j\right\}$ \text{ ;} $\quad$ $\mathcal{X}\leftarrow \mathcal{X}$ ; $ \quad \mathbf{w}^{Y} \leftarrow \mathbf{w}^{Y} - p_j \mathbf{w}^{j}$
            \IF{$|\mathcal{X}| = B_c$}
        \STATE  \textbf{break}
    \ENDIF
        \ENDFOR
        \ENDFOR
    \STATE \textbf{Return} $\mathcal{X}$
\end{algorithmic}
\end{algorithm}

\newpage
\section{Proof of \cref{theorem1}}
\label{app:proof}

Both algorithms, GGC and BGGC, for a given client $k$, for a given set of clients $\mathcal{S}$, aim to find a set of collaborators $\mathcal{X} \subseteq \mathcal{S} \cup {k}$ that maximizes the reward function $\mathcal{R}(\mathcal{X}) \triangleq - F_k^{\mathcal{V}}\left(\frac{1}{\sum_{i \in \mathcal{X}} p_i} \sum_{i \in \mathcal{X}} p_i \mathbf{w}_i\right)$. Both algorithms use the marginal gains $a$ and $b$ computed from the reward function $\mathcal{R}(\cdot)$ and add a client with the same probability function $p = \frac{a}{a + b}$. Both BGGC and GGC initialize $\mathcal{X}$ as $\{k\}$ and $\mathcal{Y}$ as $\mathcal{S} \cup \{k\}$. Both algorithms go through the potential collaborators in $\mathcal{S}$ and decide to add or remove sequentially. 

Assuming both GGC and BGGC apply some seeded sorting function to set $\mathcal{S}$ and follow the shuffled order, to show that both methods yield exactly the same output with seeded randomness, it only needs to be verified that the computed $p$ is the same for GGC and BGGC for each client. This requires showing that the reward computation yields exactly the same outcome.

In the following, we show that GGC and BGGC compute the same reward function of the four considered sets in every decision round, which are $\mathcal{X}$, $\mathcal{X} \cup \{k\}$, $\mathcal{Y}$ and $\mathcal{Y} \setminus \{k\}$, leading to the same value for the probability function $p$, hence the same decision. 

\textbf{Reward of $\mathcal{Y}$ computation}

GGC computes $\mathcal{R}(\mathcal{Y})$ as follows:
\begin{align}
\mathcal{R}(\mathcal{Y}) = - F_k^{\mathcal{V}}\left(\frac{1}{\sum_{i \in \mathcal{Y}} p_i} \sum_{i \in \mathcal{Y}} p_i \mathbf{w}_i\right),
\end{align}
which implicitly assumes having access to all the weights of clients in  $\mathcal{Y}$, possibly requiring communication with all these clients as well as the storage of their models.

In contrast, BGGC, does not assume the possibility of communicating or storing more models than the expected budget, hence in its first loop (lines 2-6) iterates through batches of clients, summing their weighted models into $\textbf{w}^Y$. This effectively pre-computes the sum for the entire client set $\mathcal{Y}= \mathcal{S}\cup \{k\}$:
 \begin{align}
   \mathbf{w}^Y &= p_k \mathbf{w}_k + \sum_{s\in \lceil\frac{n}{\mathcal{B}_c}\rceil} \sum_{j \in \mathcal{B}_s} p_j \textbf{w}_j
 &=  p_k \mathbf{w}_k + \sum_{j \in \mathcal{S}} p_j \textbf{w}_j  
  &=  \sum_{j \in \mathcal{S} \cup \{k\}} p_j \textbf{w}_j 
  &=  \sum_{j \in \mathcal{Y}} p_j \textbf{w}_j,
 \end{align}
where $\mathcal{B}_s$ denotes a batch of clients received in the s-th iteration. This weighted summation clearly ends up summing all the weights of all the clients in $\mathcal{S}$.

This summation is adaptive to the change in $\mathcal{Y}$, as shown in line 23, where whenever a client $j$ is removed, its weighted weights $(p_j \mathbf{w}^j)$ are removed from the weighted sum ($\mathbf{w}^{Y} - p_j \mathbf{w}^{j}$). Therefore, $\mathbf{w}^Y$ always represents the weighted sum of the model's weights in the set $\mathcal{Y}$. Hence, in every decision step, it is always the case that:
 \begin{align}
 \label{eq:sum_is_y}
   \mathbf{w}^Y 
  &=  \sum_{j \in \mathcal{Y}} p_j \textbf{w}_j . 
 \end{align}

BGGC computes $\mathcal{R}(\mathcal{Y})$, in line 17, as follows:
\begin{align*}
\mathcal{R}(\mathcal{Y}) = -F_k^{\mathcal{V}}\left(\frac{\mathbf{w}^{Y}}{\sum_{i \in \mathcal{Y}} p_i} \right).
\end{align*}
Replacing by $\mathbf{w}^Y$ recovers the same reward computation of GGC. \\

\textbf{Reward of $\mathcal{Y}\setminus\{k\}$ computation}

GGC computes $\mathcal{R}(\mathcal{Y}\setminus\{j\})$ as follows:
\begin{align*}
\mathcal{R}(\mathcal{Y}\setminus\{j\}) = - F_k^{\mathcal{V}}\left(\frac{1}{\sum_{i \in \mathcal{Y}\setminus\{j\}} p_i} \sum_{i \in \mathcal{Y}\setminus\{j\}} p_i \mathbf{w}_i\right),
\end{align*}
which implicitly assumes having access to all the weights of clients in  $\mathcal{Y}$.

BGGC computes $\mathcal{R}(\mathcal{Y})$, in line 16, as follows:
\begin{align*}
\mathcal{R}(\mathcal{Y}\setminus \left\{j\right\}) &= -F_k^{\mathcal{V}}\left(\frac{\mathbf{w}^{Y} - p_j \mathbf{w}^{j}}{- p_j + \sum_{i \in \mathcal{Y} } p_i} \right) 
= -F_k^{\mathcal{V}}\left(\frac{\mathbf{w}^{Y} - p_j \mathbf{w}^{j}}{\sum_{i \in \mathcal{Y}\setminus\{j\} } p_i} \right).
\end{align*}
Replacing by $\mathbf{w}^Y$, using \cref{eq:sum_is_y} recovers the same reward computation of GGC.

\textbf{Reward of $\mathcal{X}$ computation}

GGC computes $\mathcal{R}(\mathcal{X})$ as follows:
\begin{align*}
\mathcal{R}(\mathcal{X}) = - F_k^{\mathcal{V}}\left(\frac{1}{\sum_{i \in \mathcal{X}} p_i} \sum_{i \in \mathcal{X}} p_i \mathbf{w}_i\right).
\end{align*}

BGGC initializes $\mathcal{X} = \{k\}$ in the same way as GGC. Moreover, initialized  $\mathbf{w}^{X}\leftarrow p_k \mathbf{w}^k$. Therefore, in the first iteration, $\mathbf{w}^{X}$ represents exactly the weighted weight of client $k$, representing the sum of that single element.

This summation is adaptive to the change in $\mathcal{X}$, as shown in line 21, where whenever a client $j$ is added, its weighted weights $(p_j \mathbf{w}^j)$ are added to the weighted sum ($\mathbf{w}^{X} + p_j \mathbf{w}^{j}$). Therefore, $\mathbf{w}^X$ always represents the weighted sum of the model's weights in the set $\mathcal{X}$. Hence, in every decision step, it is always the case that:
 \begin{align}
 \label{eq:sum_is_X}
   \mathbf{w}^X
  &=  \sum_{j \in \mathcal{X}} p_j \textbf{w}_j . 
 \end{align}

BGGC computes $\mathcal{R}(\mathcal{Y})$, in line 14, as follows:
\begin{align*}
\mathcal{R}(\mathcal{X}) = -F_k^{\mathcal{V}}\left(\frac{\mathbf{w}^{X}}{\sum_{i \in \mathcal{X}} p_i}\right).
\end{align*}

Replacing by $\mathbf{w}^X$ recovers the same reward computation of GGC.

\textbf{Reward of $\mathcal{X}\cup\{k\}$ computation}

GGC computes $\mathcal{R}(\mathcal{X}\cup \left\{j\right\})$ as follows:
\begin{align*}
\mathcal{R}(\mathcal{X}\cup \left\{j\right\}) = - F_k^{\mathcal{V}}\left(\frac{1}{\sum_{i \in \mathcal{X} \cup \{j\}} p_i} \sum_{i \in \mathcal{X} \cup \{j\}} p_i \mathbf{w}_i\right)
\end{align*}

BGGC computes $\mathcal{R}(\mathcal{Y})$, in line 15, as follows:
\begin{align*}
\mathcal{R}(\mathcal{X} \cup \left\{j\right\}) = -F_k^{\mathcal{V}}\left(\frac{\mathbf{w}^{X} + p_j \mathbf{w}^{j}}{p_j + \sum_{i \in \mathcal{X} } p_i} \right)
= -F_k^{\mathcal{V}}\left(\frac{\mathbf{w}^{X} + p_j \mathbf{w}^{j}}{\sum_{i \in \mathcal{X}\cup\{j\} } p_i} \right)
\end{align*}

Replacing by $\mathbf{w}^X$ using \cref{eq:sum_is_X} recovers the same reward computation of GGC.

\section{Proof of \cref{proposition1}}
\label{app:proof_proposition}

Assume there exists a collaboration graph $\mathcal{C}_c$, which is optimal within a restricting subset $\mathcal{P}$, i.e.,

$$
\begin{aligned}
 \mathcal{C}_c \in \argmin_{\substack{\mathbf{\mathbf{w}} = \{\mathbf{w}_i \in \mathbb{R}^{d} \}_{i=1}^N\\
\mathbf{\mathcal{C}} = \{\mathcal{C}_i \in 2^{[N] \setminus \{i\}};\textbf{  } |\mathcal{C}_i| \leq B_c\}_{i=1}^N\\ 
\mathbf{\mathcal{C}} \in \mathcal{P}
 }} 
\left\{F(\mathbf{\mathbf{w}},\mathbf{\mathcal{C}}) \triangleq \sum_{k=1}^N p_k F_k\left(\mathbf{w}_k, \mathcal{C}_k\right)\right\}\\
\end{aligned}
$$

Solving \cref{eq:perso_collaborators_initial} yields a collaboration graph $\mathcal{C}^\star$, where

$$
\begin{aligned}
 \mathcal{C}^\star \in  \argmin_{\substack{\mathbf{\mathbf{w}} = \{\mathbf{w}_i \in \mathbb{R}^{d} \}_{i=1}^N\\
\mathbf{\mathcal{C}} = \{\mathcal{C}_i \in 2^{[N] \setminus \{i\}};\textbf{  } |\mathcal{C}_i| \leq B_c\}_{i=1}^N,\\ 
 }} 
\left\{F(\mathbf{\mathbf{w}},\mathbf{\mathcal{C}}) \triangleq \sum_{k=1}^N p_k F_k\left(\mathbf{w}_k, \mathcal{C}_k\right)\right\}\\
\end{aligned}
$$

Therefore, it follows that
\begin{align*}
\min_{\substack{\mathbf{\mathbf{w}} = \{\mathbf{w}_i \in \mathbb{R}^{d} \}_{i=1}^N\\
 }} F(\mathbf{\mathbf{w}},\mathbf{\mathcal{C}^{\star}}) &= \min_{\substack{\mathbf{\mathbf{w}} = \{\mathbf{w}_i \in \mathbb{R}^{d} \}_{i=1}^N\\
\mathbf{\mathcal{C}} = \{\mathcal{C}_i \in 2^{[N] \setminus \{i\}};\textbf{  } |\mathcal{C}_i| \leq B_c\}_{i=1}^N,\\ 
 }} 
\left\{F(\mathbf{\mathbf{w}},\mathbf{\mathcal{C}}) \triangleq \sum_{k=1}^N p_k F_k\left(\mathbf{w}_k, \mathcal{C}_k\right)\right\} \\
&\leq \min_{\substack{\mathbf{\mathbf{w}} = \{\mathbf{w}_i \in \mathbb{R}^{d} \}_{i=1}^N\\
\mathbf{\mathcal{C}} = \{\mathcal{C}_i \in 2^{[N] \setminus \{i\}};\textbf{  } |\mathcal{C}_i| \leq B_c\}_{i=1}^N\\ 
\mathbf{\mathcal{C}} \in \mathcal{P}
 }} 
\left\{F(\mathbf{\mathbf{w}},\mathbf{\mathcal{C}}) \triangleq \sum_{k=1}^N p_k F_k\left(\mathbf{w}_k, \mathcal{C}_k\right)\right\} \\
&=\min_{\substack{\mathbf{\mathbf{w}} = \{\mathbf{w}_i \in \mathbb{R}^{d} \}_{i=1}^N\\
 }} F(\mathbf{\mathbf{w}},\mathbf{\mathcal{C}_c}),
\end{align*}
which concludes the proof.

\newpage

\section{Experimental Details}
\label{app:detailed_exp}
All the experiments reported in Tables \ref{tab:sensitivity_analysis_1}, and \ref{tab:Accu_image} represent the average of three different repetitions across three different seeds. For the experiment in \cref{tab:sensitivity_analysis_2}, we report the results for seed equals to 42.  In all experiments, we preserve the best model based on the validation dataset, and the reported test results are obtained by performing inference on this best validation model.

\subsection{Datasets}
In our experiments, we utilize various datasets following existing literature on personalized FL, as highlighted in \cite{ditto, fedrep, marfoq2022personalized}. We conduct experiments with CIFAR10 \cite{krizhevsky2009learning}, Federated Extended MNIST (FEMNIST) \cite{caldas2018leaf}, and the CINIC10 dataset \cite{darlow2018cinic}. 

\subsection{Data heterogeneity} 
We explore varying degrees of data heterogeneity, particularly employing two distinct distribution strategies for the CIFAR10 and CINIC10 datasets. The first approach involves a pathological distribution split \cite{zhang2023fedala, mcmahan2017communication, colin2016gossip}, wherein each client exclusively receives data from three specified categories. The second approach utilizes the Dirichlet distribution \cite{yurochkin2019bayesian, wang2020federated}, where a distribution vector $\mathbf{q}_{c}$ is drawn from $Dir_k(\alpha)$ for each category $c$. Subsequently, the proportion $\mathbf{q}_{c,i}$ of data samples from category $c$ is allocated to client $i$. Moreover, for the FEMNIST dataset, we consider the natural Non-IID split provided in the Leaf framework \cite{caldas2018leaf} where each writer corresponds to a client and we add more degrees of heterogeneity by having each client missing some classes.

\subsection{CIFAR10 benchmark}
\subsubsection{Distribution}
CIFAR10 is a vision dataset comprising 50,000 training images and 10,000 testing images. We split the training data into 20\% of the validation dataset and 80\% of the training dataset. Furthermore, we split the testing data among clients in such a way the local test data follows the distribution of the training data. During the training, we save the best local models on the validation dataset, and we make inferences afterward using the local test data and the best saved model. 
To simulate real-world heterogeneity, we consider two types of data heterogeneity. The first approach involves a pathological distribution split \cite{zhang2023fedala, mcmahan2017communication, colin2016gossip}, wherein each client exclusively receives data from three specified categories. In the case of the CIFAR10 dataset, we use $Patho(3)$ to denote that each client has access to only three classes out of ten. The second approach utilizes the Dirichlet distribution \cite{yurochkin2019bayesian, wang2020federated}, where a distribution vector $\mathbf{q}_{c}$ is drawn from $Dir_k(\alpha)$ for each category $c$. Subsequently, the proportion $\mathbf{q}_{c, i}$ of data samples from category $c$ is allocated to client $i$. In CIFAR10 dataset we use $Dir(0.1)$. 
\subsubsection{Model} The employed model is a simple CNN network comprising three convolutional layers and two fully connected layers. The first convolutional layer has three input channels, six output channels, and a kernel size of 5. The ReLU activation function and a 2D Maxpool Layer with a kernel size of 2 follow it. The second convolutional layer transforms an input of six channels to sixteen channels with a kernel size of 5, followed by the ReLU activation function. The first fully connected layer takes an input of size 400 and produces an output of size 120. The second layer produces an output of size 84, and the last layer has a size equal to the number of classes, which is 10.

\subsubsection{Hyperparameters}
For the preprocessing step for each client, we train 10 local epochs ($\tau_{init}=10$). During the training the number of local epochs $\tau_{train} = 5$, the number of rounds $T=100$, however as the preprocessing step corresponds to 2 rounds of training, for fairness with other methods we use only 98 rounds instead of 100 for our method. The number of clients $|\mathcal{S}_t| = 100$. The learning rate $\eta = 0.01$. For the training, we use SGD optimizer with $1e-3$ decay, 0.9 momentum, and batch size of 16. 

\subsection{FEMNIST benshmark}
\subsubsection{Distribution}
We utilize the FEMNIST dataset within the LEAF framework \cite{caldas2018leaf}. This dataset consists of training and testing sets accompanied by a client-data mapping file that partitions the data in a non-IID (non-identically distributed) manner among the clients. The dataset exhibits inherent heterogeneity due to variations in the writing styles of individual contributors. We first downloaded the full FEMNIST data from \cite{caldas2018leaf}. Additionally, we employed a file obtained from \cite{ditto} to further increase the heterogeneity in the dataset. The data files will be provided along with our code.

\subsubsection{Model}

The employed model is a simple CNN network comprising three convolutional layers and two fully connected layers. The first convolutional layer has one input channel, 4 output channels, and a kernel size of 5. It is followed by the ReLU activation function and a 2D Maxpool Layer with a kernel size of 2. The second convolutional layer transforms an input of 4 channels to 12 channels with a kernel size of 5, followed by the ReLU activation function. The first fully connected layer takes an input of size 192 and produces an output of size 120. The second layer produces an output of size 100, and the last layer has a size equal to the number of classes, which is 10.

\subsubsection{Hyperparameters}
For the preprocessing step for each client, we train 4 local epochs ($\tau_{init}=4$). During the training the number of local epochs $\tau_{train} = 2$, the number of rounds $T=50$, however as the preprocessing step corresponds to 2 rounds of training, for fairness with other methods we use only 48 rounds instead of 50 for our method. The number of clients $|\mathcal{S}_t| = 100$. The learning rate $\eta = 0.001$. For the training, we use SGD optimizer with $1e-3$ decay, 0.9 momentum, and batch size of 10.

\subsection{CINIC10 benshmark}
\subsubsection{Distribution}
 CINIC10 serves as an extension to CIFAR10, encompassing 90,000 training images, 90,000 validation images, and 90,000 test images. Initially, the training and validation sets are combined, and the data is then distributed among clients in a heterogeneous manner. Subsequently, 20\% of each partition is allocated for validation, with the remaining portion designated for training. Similar to CIFAR10, we split the test data among clients in such a way that it follows the train distribution for each client. The training data is distributed either following a Dirichlet distribution $Dir(0.1)$ or the pathological distribution $Patho(3)$, where each client has access to only three classes among the ten available. After assigning to each client which classes it will get, we distribute the class data points among clients that share the same class, following a $Dir(0.5)$ distribution. This additional step adds more degrees of heterogeneity and simulates a real-world scenario more realistically.
\subsubsection{Model}

The same model used in CIFAR10 is employed.

\subsubsection{Hyperparameters}

For the preprocessing step for each client, we train 10 local epochs ($\tau_{init}=10$). During the training the number of local epochs $\tau_{train} = 5$, the number of rounds $T=50$, however as the preprocessing step corresponds to 2 rounds of training, for fairness with other methods we use only 48 rounds instead of 50 for our method. The number of clients $|\mathcal{S}_t| = 200$. The learning rate $\eta = 0.01$. For the training, we use SGD optimizer with $1e-3$ decay, 0.9 momentum, and batch size of 16.

\subsection{Baselines}
We compared our method against eleven baselines:
\begin{itemize}
    \item Local: local training for $T$ rounds and we report the average local test accuracies of the clients in every round. 
    \item FedAvg \cite{mcmahan2017communication}
    \item FedAvg+FT: which is fedavg fine-tuning version, where we save best models on validation dataset when training FedAvg, then starting from that model we perform $2*\tau$ local epochs and report the average test accuracies. 
    \item FedProx \cite{lian2018asynchronous}: For FedProx we use $\lambda = 0.1$ for all datastes (CIFAR10, CINIC10 and FEMNIST)

    \item FedProx+FT \cite{lian2018asynchronous}: It is FedProx fine-tuning version, where we save the best models on the validation dataset when training FedProx, then starting from that model we perform $2*\tau$ local epochs and report the average test accuracies.

    \item APFL \cite{smith2017federated}: We chose the hyperparameter which they call $\tau$ in their paper to be $\tau = 1$ which means we synchronise the models every round.

    \item PerFedAvg \cite{fallah2020personalized}: We use the hyperparameter $\alpha = 0.01$ and it is fixed for all runs and all datasets

    \item Ditto \cite{ditto}: We set the hyperparameter $\lambda$ that represents the tradeoff between local and global models to 0.75, and keep it fixed for all datasets.
    \item  FedRep \cite{fedrep}

    \item kNN-Per \cite{marfoq2022personalized}: we set the hyperparameter $k_knn$ to 10 and the interpolation hyperparameter to 0.5 

    \item pFedGraph \cite{ye2023personalized}

\end{itemize}

\subsection{Compute resources}
\label{app:compute_resources}
We use an internal SLURM cluster for running our experiments.  The experiments were done on an ASUS ESC N4A-E11 server. The node has 4 A100 GPUs, an AMD EPYC 7003 series 64 core @ 3.5GHz CPU and 512GB of RAM.
We used one A100, with 2 cores, and required at most 100GB of memory for the experiments.

\section{Additional Experiments}
\subsection{Variance between accuracies of local models}
\begin{figure}[h]
\begin{minipage}{\textwidth}
\centering
\begin{tikzpicture}
  \node (img)  {\includegraphics[scale=0.35]{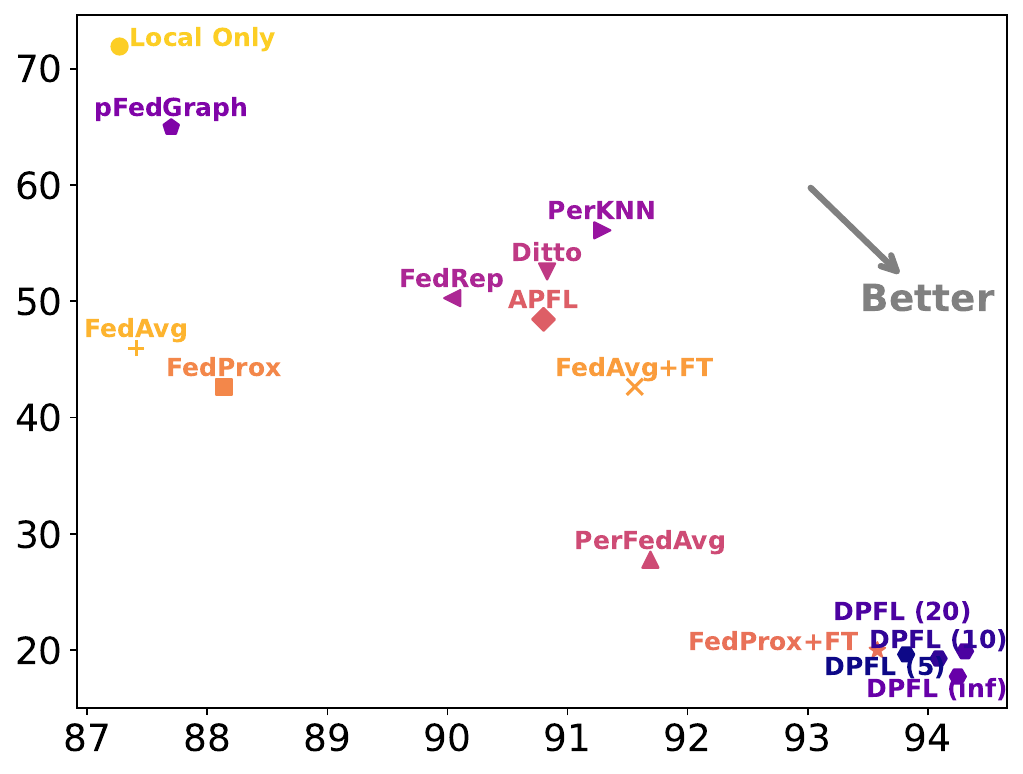}};
  \node[rotate=0cm, anchor=center,yshift=-2.5cm] {\small Average Test Accuracy};
  \node[ rotate=90, anchor=center,yshift=3.4cm] {\small Varinace};
 \end{tikzpicture}
\end{minipage}%
\caption{Comparison of our method with other personalized methods on the FEMNIST dataset in terms of variance between local models. }
\label{fig:femnist_variance}
\end{figure}

\begin{figure}[h]
\begin{minipage}{0.48\textwidth}
\centering
\begin{tikzpicture}
  \node (img)  {\includegraphics[scale=0.35]{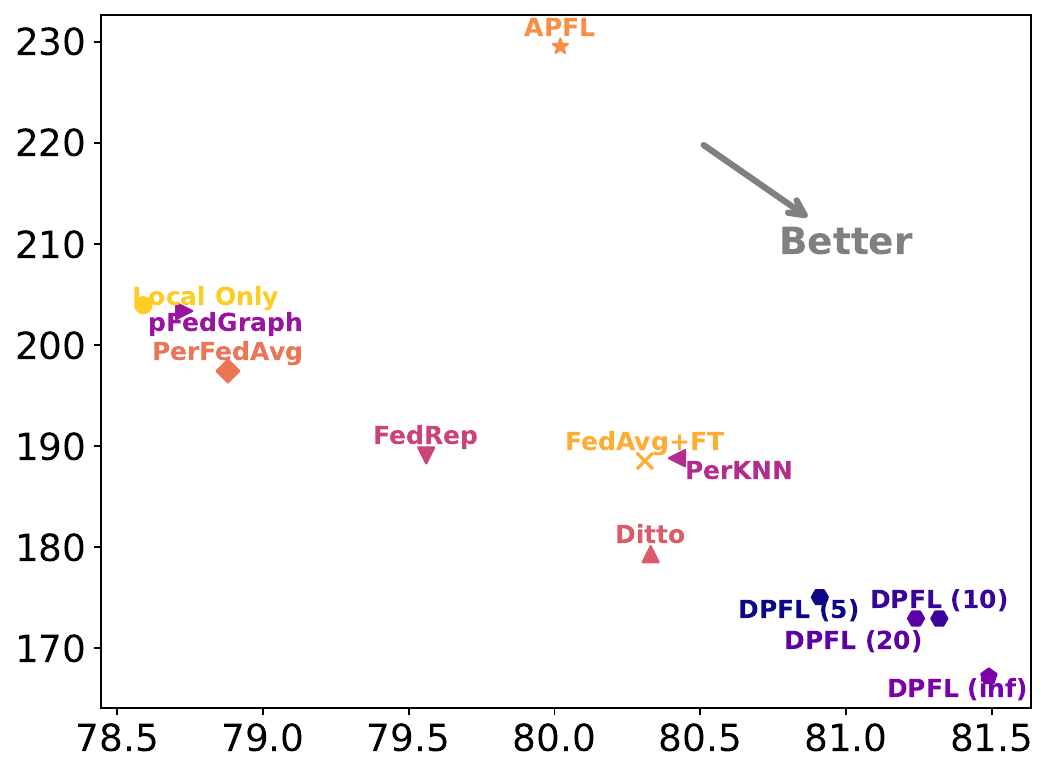}};
  \node[rotate=0cm, anchor=center,yshift=-2.5cm] {\small Average Test Accuracy};

 \end{tikzpicture}
\end{minipage}%
\begin{minipage}{0.48\textwidth}
\centering
\begin{tikzpicture}
  \node (img)  {\includegraphics[scale=0.35]{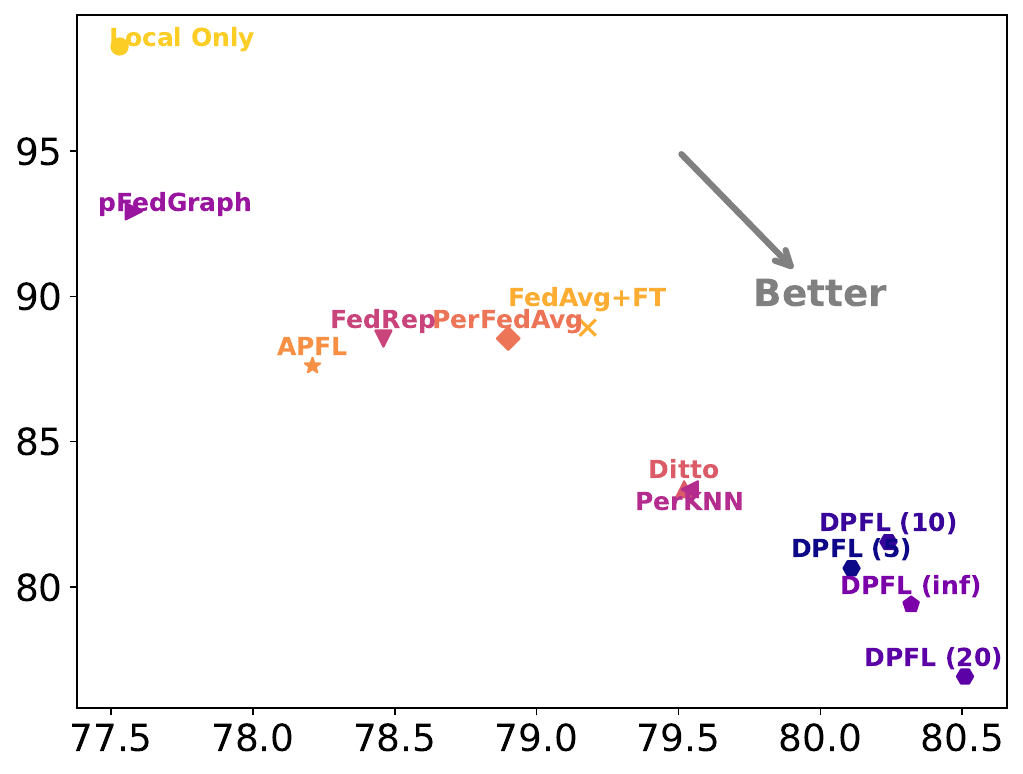}};
  \node[ rotate=0cm, anchor=center,yshift=-2.5cm] {\small Average Test Accuracy};
 \end{tikzpicture}
\end{minipage}%
\caption{Comparison of our method with other personalized methods on the CINIC10 dataset in terms of variance between local models. }
\label{fig:CINIC10_variance}
\end{figure}

As our objective is personalization, we believe that, apart from improving the overall average accuracy among clients, it is crucial to assess whether improvements are distributed across most models rather than being confined to just a few clients. To evaluate this, we examine the variance between local models. Lower variance indicates that the accuracies of local models are closer. Therefore, we consider both accuracy and variance as important metrics. In Figures \ref{fig:CIFAR10_variance}, \ref{fig:femnist_variance} and \ref{fig:CINIC10_variance}, the x-axis represents the average test accuracy, while the y-axis represents the variance between clients' models. For better visualization, in \cref{fig:CIFAR10_variance} we choose to report only methods that the average test accuracy is higher than 80\% and 78\% for $Dir(0.1)$ and $Patho(3)$ respectively, while for \cref{fig:CINIC10_variance} we show the methods having higher than 78\% and 77\% for $Dir(0.1)$ and $Patho(3)$ respectively. Furthermore, in \cref{fig:femnist_variance} we were able to show all methods. All figures show that our method (DPFL) is situated in the right-bottom corner across all variants of budget constraints. This positioning signifies that, compared to other methods, our approach achieves superior average test accuracy and lower variance between local models. 

\label{appendix:Variance}

\begin{figure}[h]
\begin{minipage}{0.49\textwidth}
\begin{tikzpicture}
      \node (img)  {\includegraphics[scale=0.35]{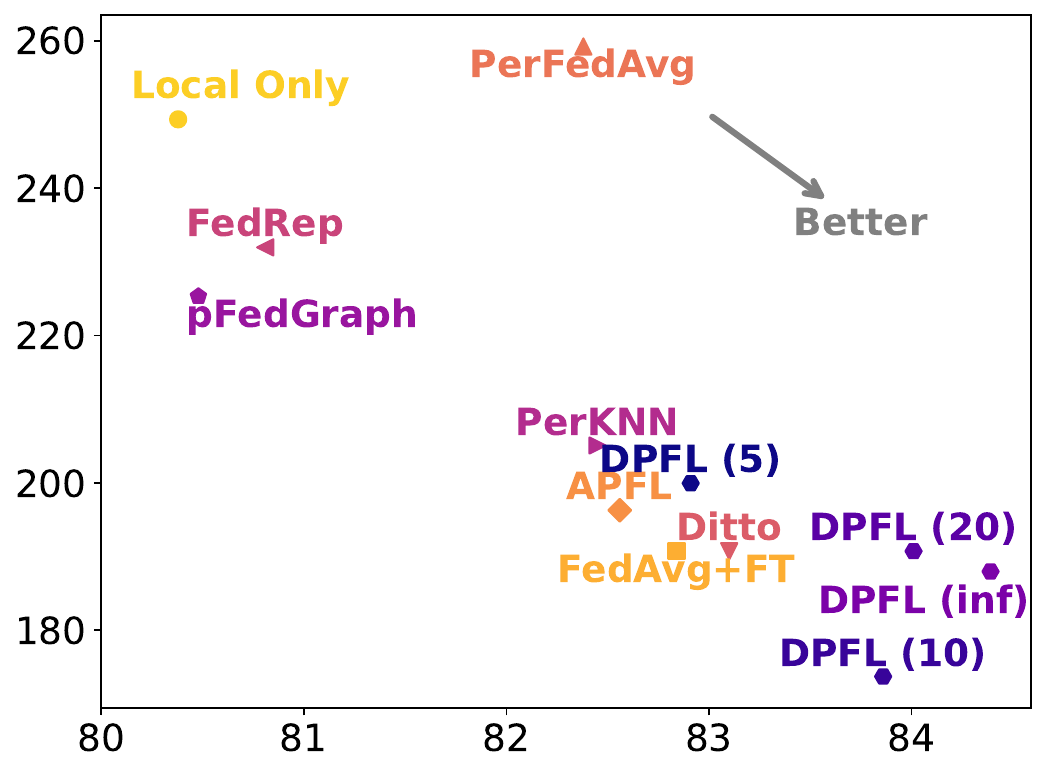}};
  \node[node distance=0cm, rotate=0cm, anchor=center,yshift=-2.5cm] {\small Test Accuracy};
  \node[node distance=0cm, rotate=90, anchor=center,yshift=3.4cm] {\small Variance};
 \end{tikzpicture}
\end{minipage}%
\begin{minipage}{0.49\textwidth}
\begin{tikzpicture}
  \node (img)  {\includegraphics[scale=0.35]{figures/CIFAR10_variance_complex_dist.pdf}};
  \node[node distance=0cm, rotate=0cm, anchor=center,yshift=-2.5cm] {\small Test Accuracy};
\end{tikzpicture}
\end{minipage}%
\caption{Variance between local models using Dir(0.1) (left) and Patho(3) (right) data splits on CIFAR10.
}
\label{fig:CIFAR10_variance_appendix}
\end{figure}

\subsection{Visualization of the collaboration graph}

\begin{figure}[t]
\centering
\includegraphics[width=\textwidth]{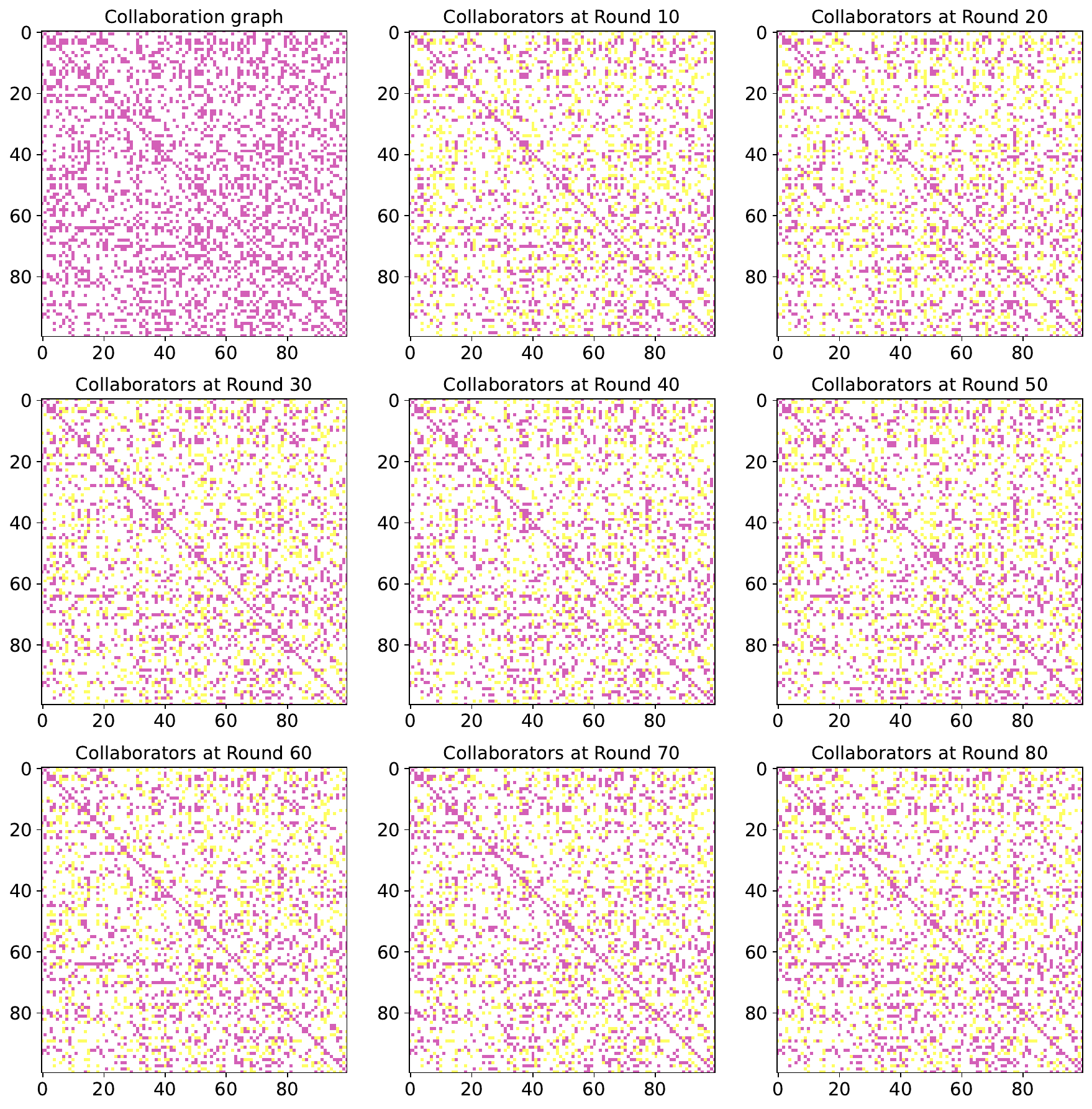}
\caption{Our collaboration graph without constraint on CIFAR10 dataset with 100 clients}
\label{fig:collaboration_graph_full_without_constraint}
\end{figure}

\begin{figure}[h]
\centering
\includegraphics[width=\textwidth]{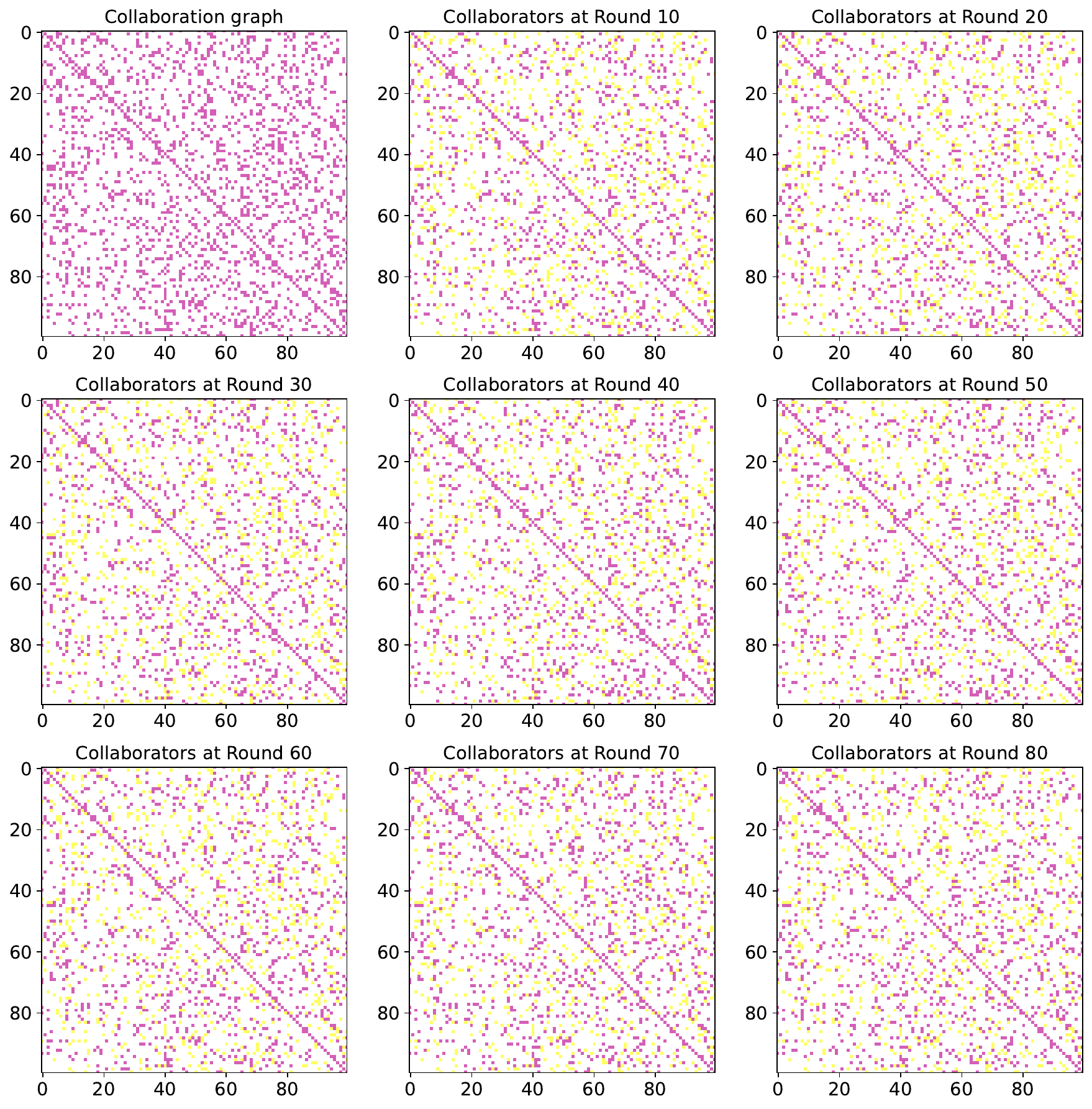}
\caption{Our collaboration graph with constraint $B_c$=20 on CIFAR10 dataset with 100 clients.}
\label{fig:collaboration_graph_full_20}
\end{figure}

\begin{figure}[h]
\centering
\includegraphics[width=\textwidth]{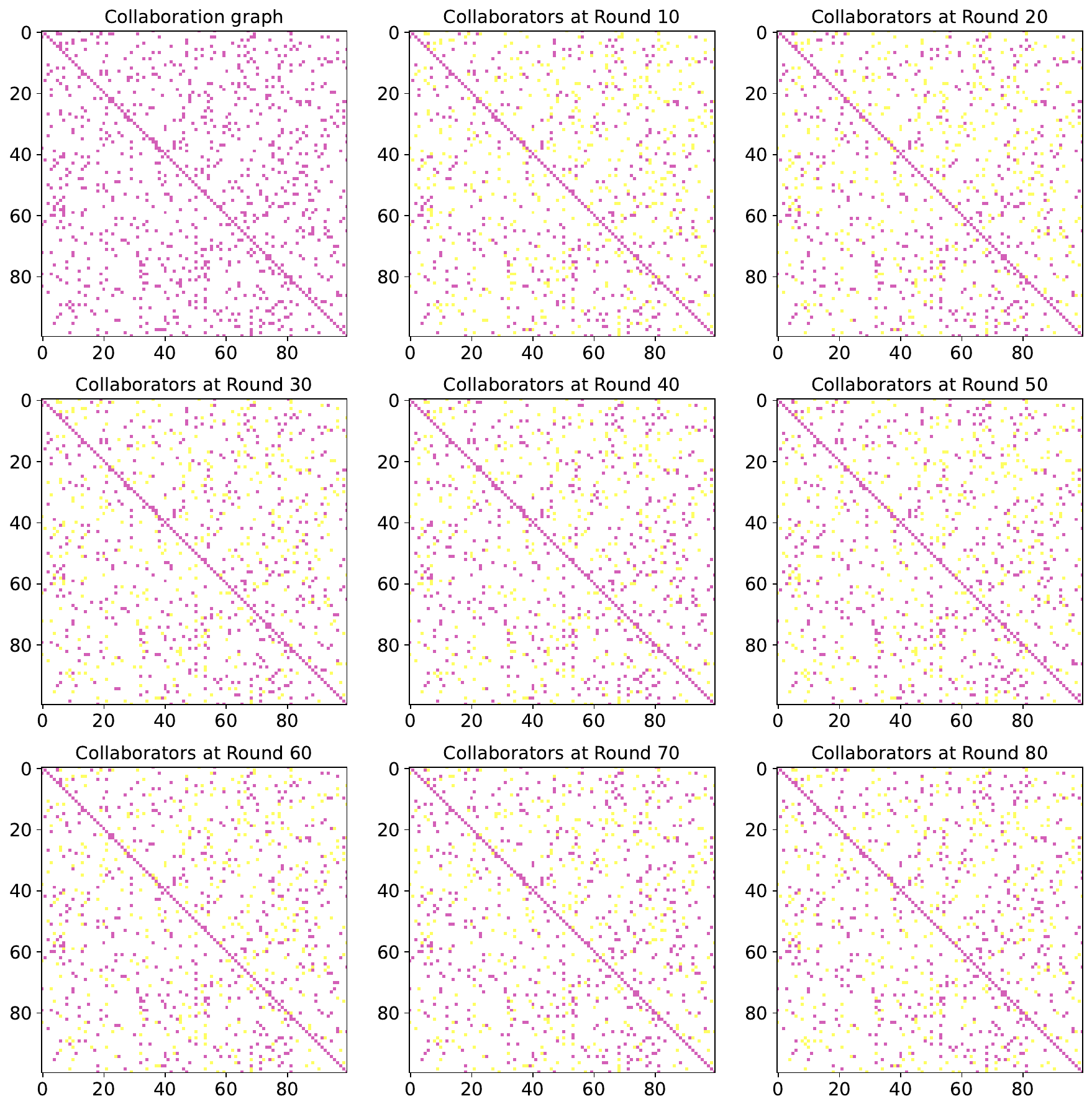}
\caption{Our collaboration graph with constraint $B_c$=10 on CIFAR10 dataset with 100 clients.}
\label{fig:collaboration_graph_full_10}
\end{figure}

\begin{figure}[h]
\centering
\includegraphics[width=\textwidth]{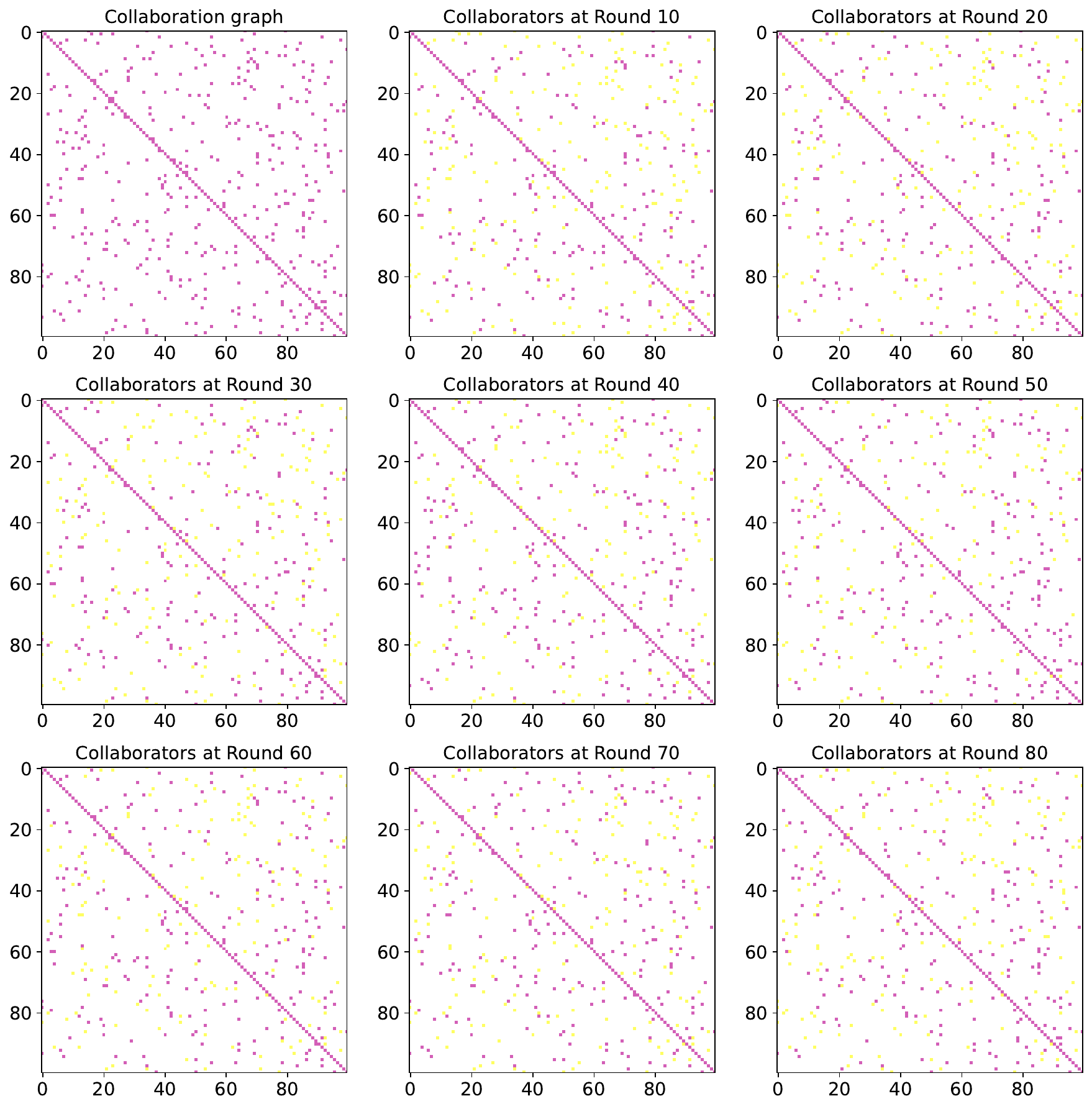}
\caption{Our collaboration graph with constraint $B_c$=5 on CIFAR10 dataset with 100 clients.}
\label{fig:collaboration_graph_full_5}
\end{figure}

We present Figures \ref{fig:collaboration_graph_full_without_constraint}, \ref{fig:collaboration_graph_full_20}, \ref{fig:collaboration_graph_full_10} and \ref{fig:collaboration_graph_full_5}, showing our initial collaboration graph on the top left and the clients considered for collaboration every 10 round from 0 to 80 for cases without budget constraint, with $B_c = 20$, $B_c = 10$, and with $B_c = 5$ respectively. The diagonal indicates that every client always ``collaborates'' with itself.
To illustrate the graph evolution, the plots display collaborative links in two colors: in pink, are the clients that are selected for collaboration in that round; in yellow, are the clients that were identified during the preprocessing step but are currently not chosen.
The union of pink and yellow clients corresponds to the initial collaboration graph. 
The figures highlight that the initial collaboration graph is denser compared to the actually used clients for aggregation in later rounds. This is expected since the graph is constructed as a preprocessing step, and at this stage, model weights have not yet converged. Therefore, broader collaboration can be beneficial. However, as training progresses, it is natural to expect that each client will benefit primarily from collaborating with clients having similar data distributions, thus leading to a sparser collaboration graph. Another contributing factor is that, in the preprocessing step, the decision to select a specific client for collaboration is made from a pool of 100 clients, making it more challenging than in later rounds where the decision is drawn from a smaller pool denoted as $\Omega_{k}$ for client $k$. Another observation is that in all cases, from round 60 onward, the graph remains almost unchanged. This is attributed to the weights nearing convergence, resulting in a less random selection of collaborators. The diversity of the graph at early rounds proves that we shouldn't remove an edge between two nodes if that node hasn't been selected, as that decision could change in later rounds.

\label{appendix:Col_graph}

\newpage
\newpage
\newpage
\newpage

\subsection{Behavior of the collaboration graph under two groups of clients}
\begin{figure}[H]
\centering
\includegraphics[width=\textwidth]{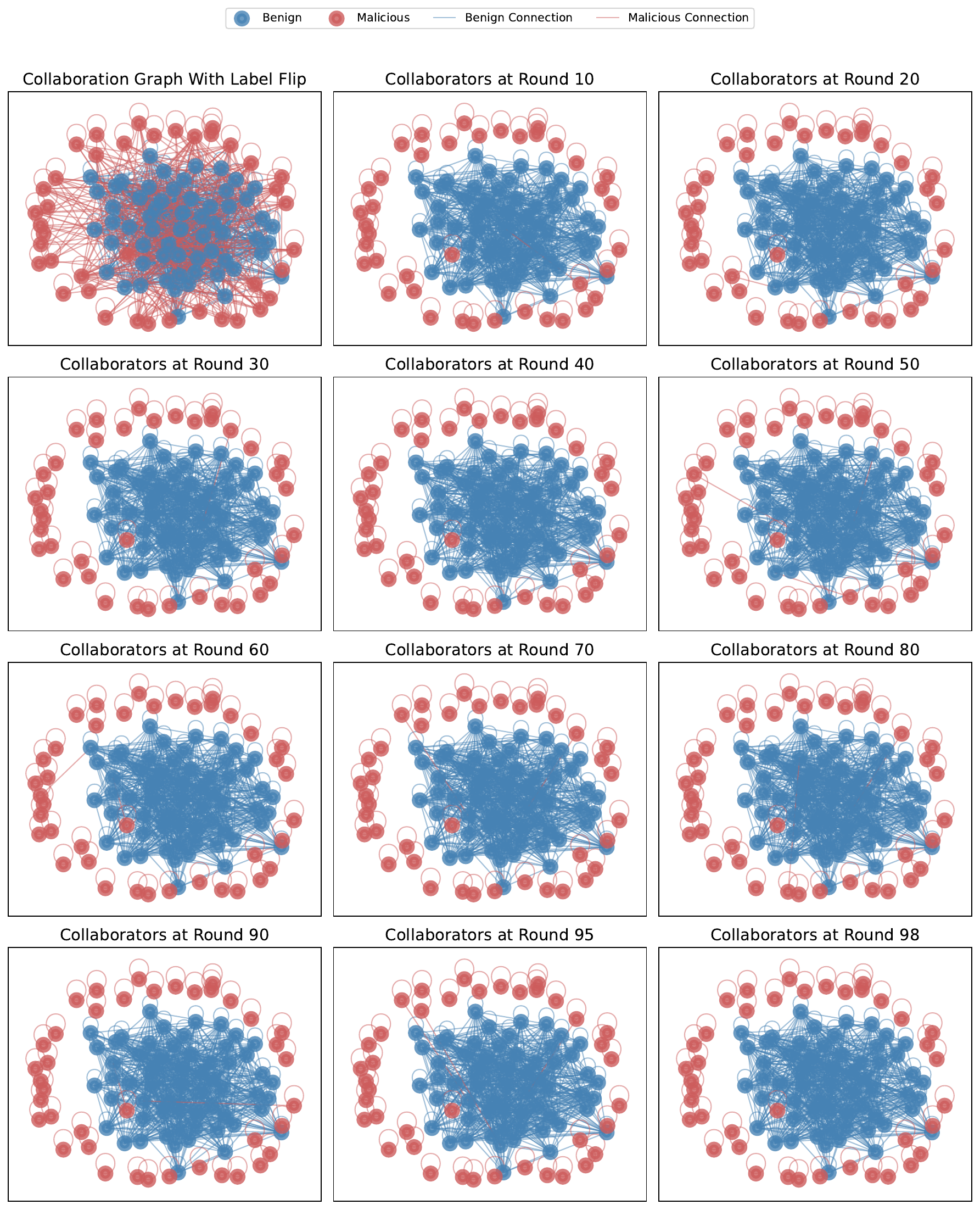}
\caption{Our collaboration graph without Greedy. 40\% of clients have flipped labels (Malicious), while 60\% have original labels (Benign). Malicious clients don't execute \cref{alg:GGC}; instead, they consistently send their local model to Benign clients.}
\label{fig:full_graph_attack_without_greedy}
\end{figure}

\begin{figure}[H]
\centering
\includegraphics[width=\textwidth]{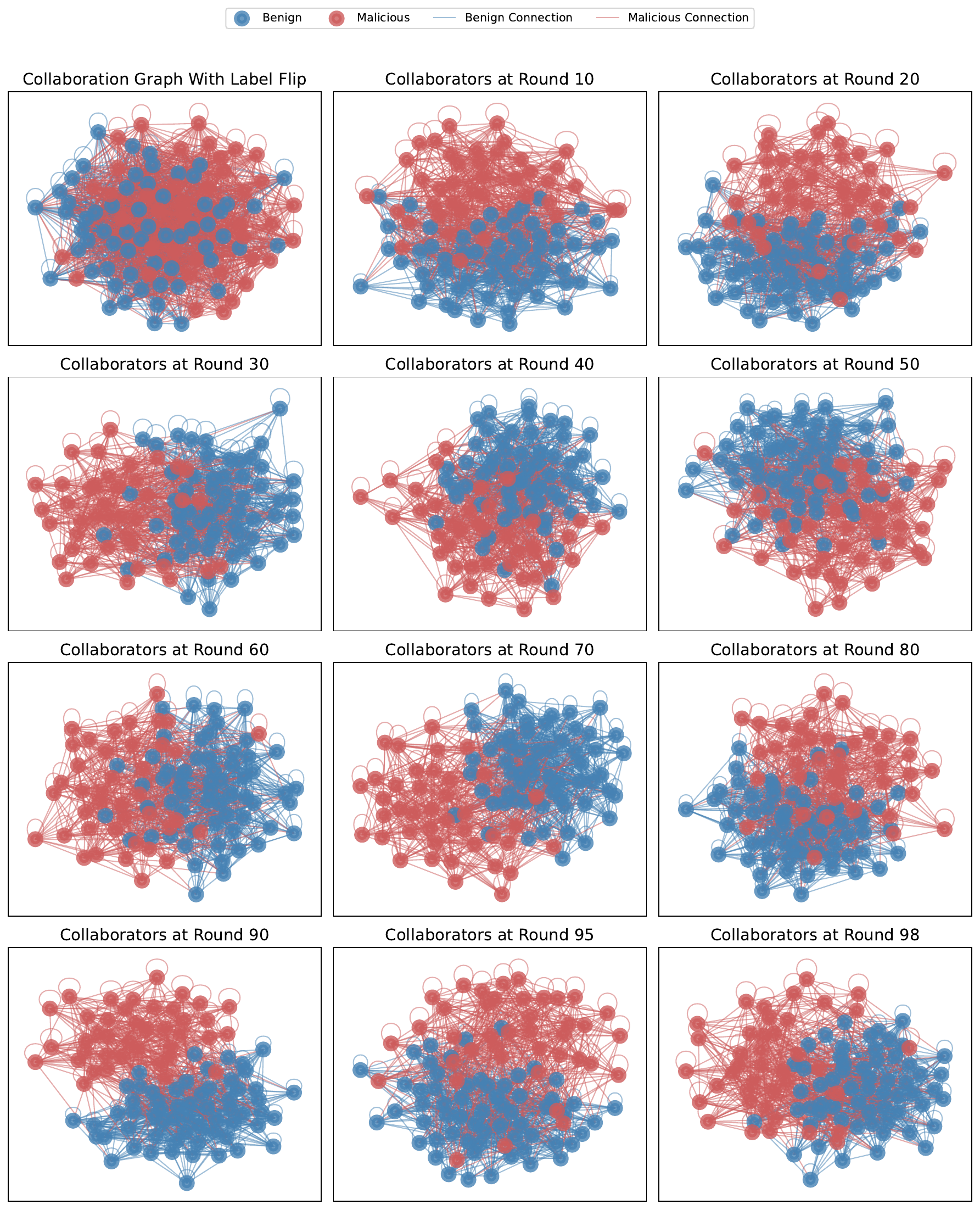}
\caption{Our collaboration graph with Greedy. 40\% of clients have flipped labels (Malicious), while 60\% have original labels (Benign). Malicious execute \cref{alg:GGC}.}
\label{fig:full_graph_attack_with_greedy}
\end{figure}

\begin{figure}[H]
\centering
\includegraphics[scale=0.3]{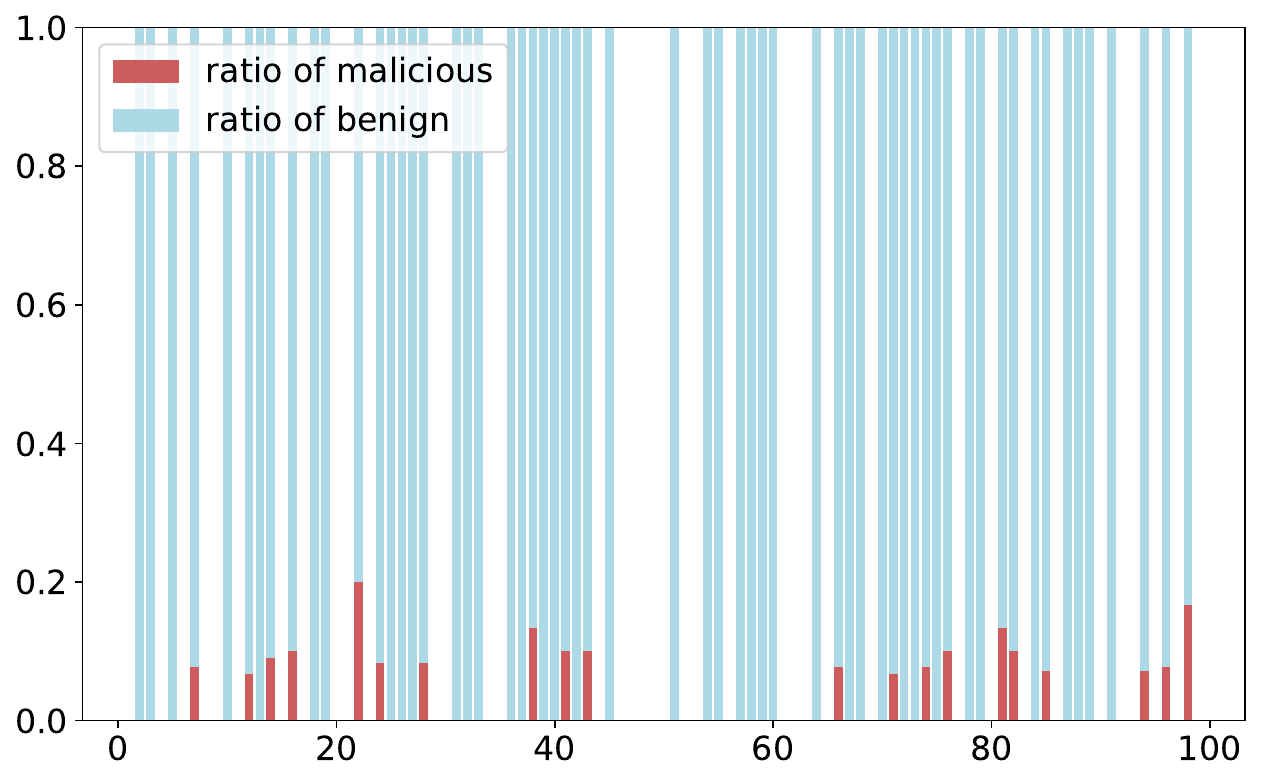}
\caption{Visualization of the ratio of benign clients collaborating with malicious vs with benign, in the case that malicious runs the greedy algorithm}
\label{fig:histo}
\end{figure}

To better visualize and have more explainability of the behavior of our algorithm, we ran an experiment using CIFAR10 dataset and 100 clients, where 40 among them had their label flipped using the same permutation and 60 had their true labels. It's important to note that our objective wasn't to create an attack; rather, we aimed to delineate the behavior of our algorithm, acknowledging that while it may exhibit robustness characteristics, the study of robustness falls outside the scope of this paper. In this experiment, we effectively created two distinct groups of clients. The first experiment, illustrated in \cref{fig:malicious_benign_without_greedy}, involved flipped clients who did not execute the greedy algorithm (\cref{alg:GGC}). Instead, they consistently maintained their models (a stronger attack strategy). As expected, the collaboration graph displayed numerous edges with these flipped clients initially. This outcome aligns with the inherent randomness in the weights during the preprocessing step, making it challenging to select between clients. However, as the rounds progressed, we observed a notable trend: black clients increasingly avoided selecting the red ones until they ultimately ceased choosing them altogether. This evolution is depicted in  \cref{fig:full_graph_attack_without_greedy}, where the complete graph evolution is visualized every 10 rounds. In the second experiment (\cref{fig:malicious_benign_with_greedy}), even the red clients executed the greedy algorithm, resulting in their selection from the black clients initially. Consequently, their models became regularized towards the black ones. Despite this behavior in the collaboration graph, we observed that as the rounds progressed, the clients were almost segregated into two subgroups (red and black), with very few links between them serving as a form of regularization (full graph is in \cref{fig:full_graph_attack_with_greedy}). We visualized the percentage of connection with malicious clients for a benign client, as presented in \cref{fig:histo}, and found it to be very small.

\newpage
$\textbf{ }$
\newpage
$\textbf{ }$
\newpage
$\textbf{ }$
\newpage
$\textbf{ }$

\newpage
$\textbf{ }$

\newpage
$\textbf{ }$

\newpage
$\textbf{ }$
\begin{figure}[H]
\centering
\includegraphics[scale=0.5]{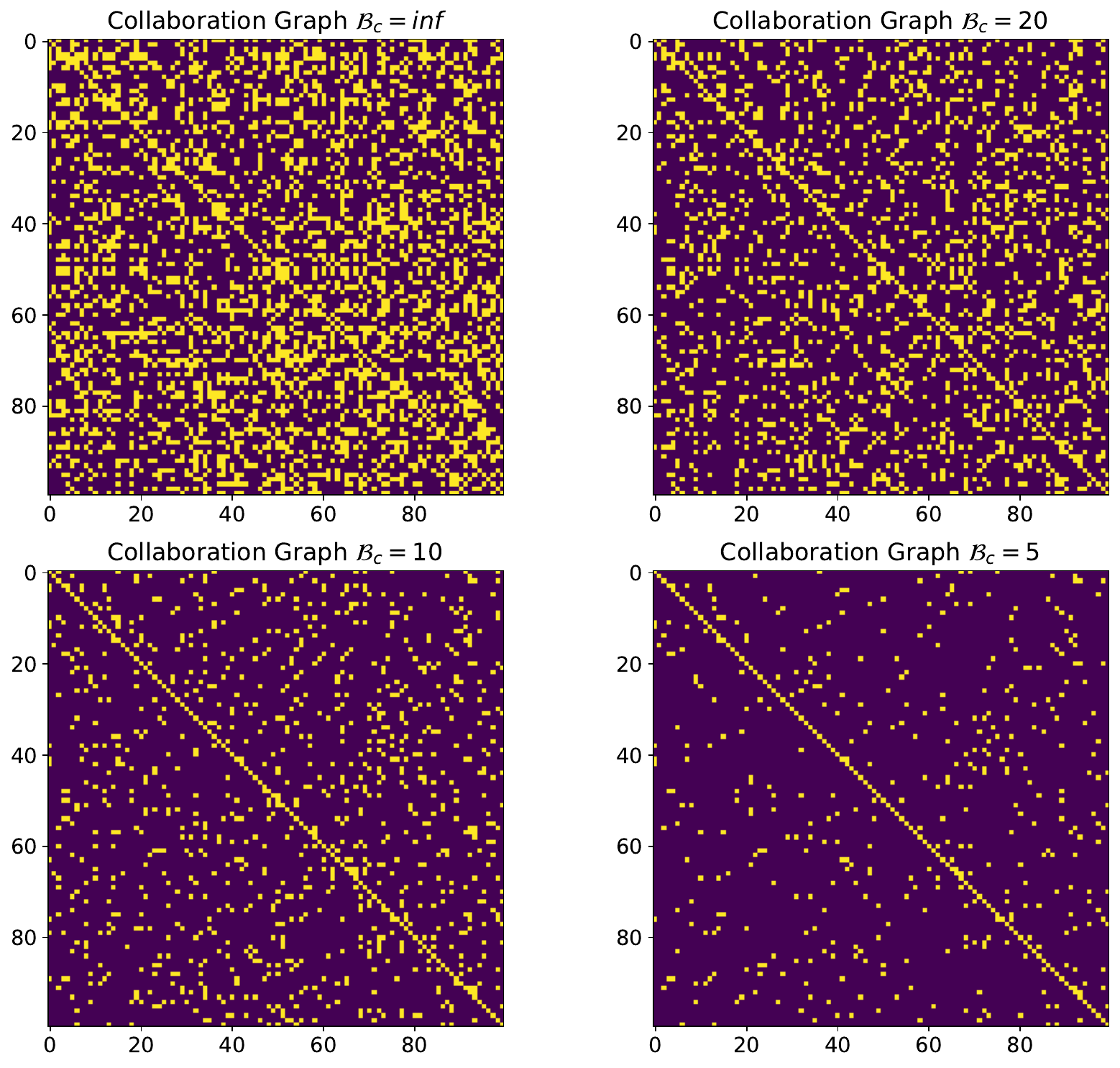}
\caption{Our collaboration graph on CIFAR10 dataset with 100 clients}
\label{fig:collaboration_graph_full_4_constraints_in_one}
\end{figure}

\subsection{Measuring Asymmetry of the collaboration graph}
\label{app:Asymmetry}

For the experiments conducted using the CIFAR10 dataset, we analyzed the percentage of asymmetry in the graph across different rounds for three budget settings $\mathcal{B}_c= inf$, $\mathcal{B}_c= 20$ and $\mathcal{B}_c= 10$ respectively.We observed that the asymmetry percentage is consistently higher in round 0 compared to subsequent rounds. This aligns with the expectation that the greedy decision-making process during preprocessing exhibits more randomness due to two factors, first, the model weights still didn't capture enough of the data structure, and second as the decision of choosing or removing a client is made from bigger pool, it makes it harder to decide. Additionally, across all budget settings, the asymmetry percentage appears to stabilize around a specific value, exhibiting fluctuations, starting from round 40 onwards.

\begin{figure*}[H]
\begin{minipage}{0.33\textwidth}
\vspace{-0.2cm}
\begin{tikzpicture}
  \node (img)  {\includegraphics[scale=0.27]{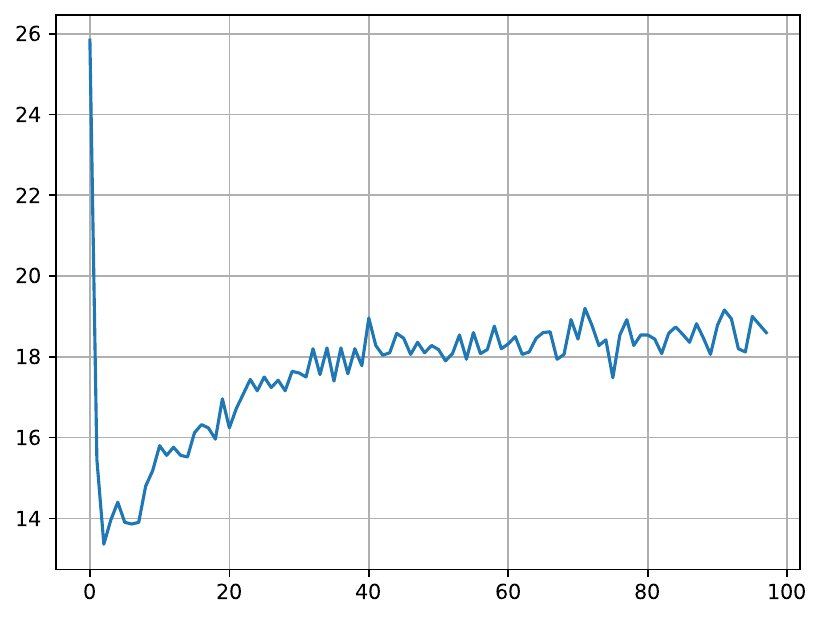}};
  \node[rotate=0cm, anchor=center,yshift=2.3cm] {Round};
  \node[rotate=90, anchor=center,yshift=3.0cm] {Asymmetry percentage};
 \end{tikzpicture}
\end{minipage}%
\begin{minipage}{0.33\textwidth}
\vspace{-0.2cm}
\begin{tikzpicture}
  \node (img)  {\includegraphics[scale=0.3]{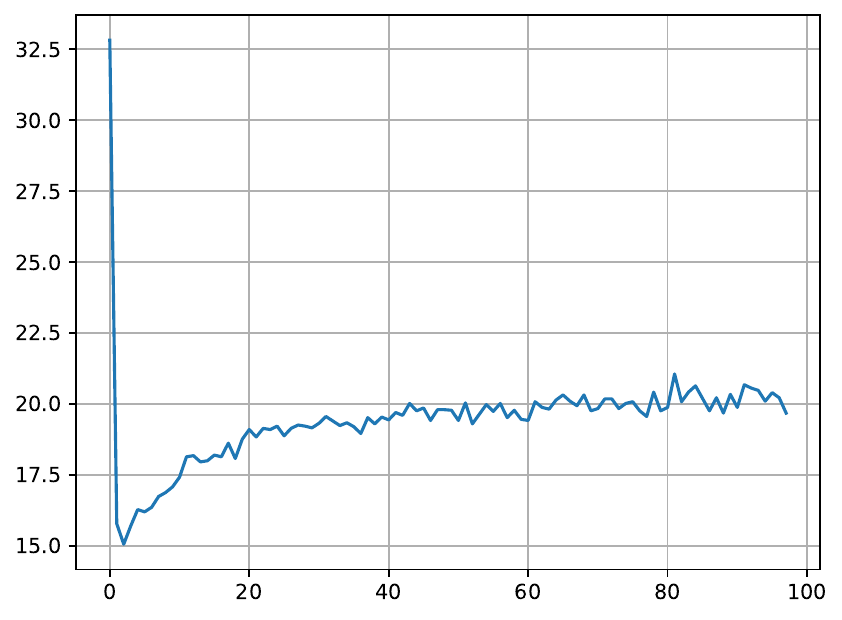}};
  \node[rotate=0cm, anchor=center,yshift=2.3cm] {Round};
\end{tikzpicture}
\end{minipage}%
\begin{minipage}{0.33\textwidth}
\vspace{-0.2cm}
\begin{tikzpicture}
  \node (img)  {\includegraphics[scale=0.3]{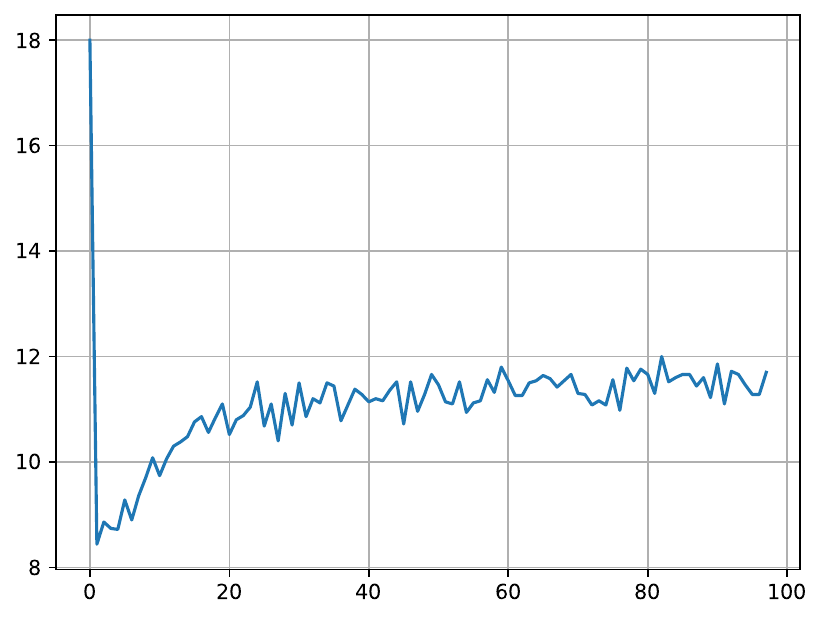}};
  \node[rotate=0cm, anchor=center,yshift=2.3cm] {Round};
\end{tikzpicture}
\end{minipage}%
\caption{Asymmetry percentage of our collaboration graph over the rounds on CIFAR10 dataset, left plot without cardinality constraint, middle $B_c = 20$, and right $B_c = 10$}
\label{fig:fedprox_femnist}
\vspace{-1em}
\end{figure*}
\end{document}